\documentclass{applemlr}

\usepackage{amsmath}
\usepackage{enumerate}
\usepackage{algorithm}
\usepackage{algpseudocode}
\usepackage{amsfonts}
\usepackage{amsthm}
\usepackage{cleveref}
\usepackage{diagbox}
\usepackage{colortbl}
\usepackage{amssymb}
\usepackage{xspace}
\usepackage{wrapfig}
\usepackage{adjustbox}
\usepackage{tabularx}
\usepackage{booktabs}
\usepackage{mathtools}
\usepackage{tikz}
\usepackage{enumitem}
\usepackage{silence}
\usepackage{dsfont}
\usepackage[table]{xcolor}
\usepackage[dvipsnames]{xcolor}
\usepackage{multirow}
\usepackage{makecell}
\usepackage{xfakebold}
\usepackage{microtype}
\usepackage{url}
\usepackage{ulem}
\usepackage{graphicx}
\usepackage{array}
\usepackage{mdframed}
\usepackage{arydshln}

\usepackage{amsmath,amsfonts,bm}

\def\eqref#1{equation~\ref{#1}}

\def\1{\bm{1}}

\DeclareMathAlphabet{\mathsfit}{\encodingdefault}{\sfdefault}{m}{sl}
\SetMathAlphabet{\mathsfit}{bold}{\encodingdefault}{\sfdefault}{bx}{n}

\usepackage{booktabs}

\definecolor{textgray}{HTML}{6E6E73}
\usetikzlibrary{positioning, calc}
\usetikzlibrary{decorations.pathmorphing}

\makeatletter
\patchcmd{\wrong@fontshape}{\@gobbletwo}{}{}{}
\makeatother
\numberwithin{equation}{section}
\makeatletter
\AtBeginDocument{
  \urlstyle{sf}
  
}
\makeatother

\definecolor{light}{RGB}{125, 125, 125}
\crefname{tcb@cnt@pbox}{code}{code}
\Crefname{tcb@cnt@pbox}{Code}{Code}
\crefname{assumption}{assumption}{assumption}
\Crefname{assumption}{Assumption}{Assumptions}

\newtcolorbox[auto counter]{pbox}[2][]{
  colback=white,
  title=Code~\thetcbcounter: #2,
  #1,fonttitle=\sffamily,
  fontupper=\sffamily,
  arc=2pt,
  colframe=bgcolor,
  coltitle=fgcolor,
  colbacktitle=bgcolor,
  toptitle=0.25cm,
  bottomtitle=0.125cm
}

\makeatletter
\newcommand\applefootnote[1]{%
  \begingroup
  \renewcommand\thefootnote{}%
  \renewcommand\@makefntext[1]{\noindent##1}%
  \footnote{#1}%
  \addtocounter{footnote}{-1}%
  \endgroup
}
\makeatother

\definecolor{cverbbg}{gray}{0.90}

\renewenvironment{quote}
  {\list{}{\rightmargin=0.4cm \leftmargin=0.4cm \small}%
   \item\relax}
  {\endlist}

\definecolor{langDE}{HTML}{b8d4ec}
\definecolor{langDElight}{HTML}{e8f2fa}
\definecolor{langFR}{HTML}{d8b8d8}
\definecolor{langFRlight}{HTML}{f2e8f2}
\definecolor{langHI}{HTML}{d8d8a8}
\definecolor{langHIlight}{HTML}{f2f2e4}
\definecolor{langZH}{HTML}{a8d8c0}
\definecolor{langZHlight}{HTML}{e4f2ec}
\definecolor{langDEvlight}{HTML}{f4f8fc}
\definecolor{langFRvlight}{HTML}{f9f4f9}
\definecolor{langHIvlight}{HTML}{f9f9f2}
\definecolor{langZHvlight}{HTML}{f2f9f6}

\definecolor{langDEavg}{HTML}{7aabcf}
\definecolor{langDElightavg}{HTML}{b8d4ec}
\definecolor{langDEvlightavg}{HTML}{d0e6f4}
\definecolor{langFRavg}{HTML}{b888b8}
\definecolor{langFRlightavg}{HTML}{d0b0d0}
\definecolor{langFRvlightavg}{HTML}{e4d0e4}
\definecolor{langZHavg}{HTML}{70b890}
\definecolor{langZHlightavg}{HTML}{a8d8c0}
\definecolor{langZHvlightavg}{HTML}{c8e8d8}
\definecolor{langHIavg}{HTML}{b8b868}
\definecolor{langHIlightavg}{HTML}{d0d0a0}
\definecolor{langHIvlightavg}{HTML}{e4e4c0}

\title{Multilingual Knowledge Transfer under Data Constraints via Lexical Interventions}

\author[1,2]{Anastasiia Sedova}
\author[1 \ddagger]{Natalie Schluter}
\author[1 \ddagger]{Skyler Seto}
\author[1 \ddagger]{Maartje ter Hoeve}

\affiliation[1]{Apple}
\affiliation[2]{ITU}

\contribution[\dagger]{Shared senior authorship}

\abstract{
Cross-lingual knowledge transfer is critical for building high-performing multilingual language models for languages with insufficient training data.
When target language data is scarce, the knowledge required for many downstream tasks involving scientific reasoning, commonsense inference, and world knowledge must be acquired primarily from the high-resource language, making effective knowledge transfer essential.
Existing methods for improving such cross-lingual knowledge transfer require large amounts of parallel data, translation systems, auxiliary models, or additional training stages that are largely unavailable for many languages.
We propose \texttt{\textbf{LINK}} -- a data-level intervention method that improves knowledge transfer during model pretraining through lexical substitutions in high-resource part of pretraining data using bilingual vocabularies.
For a given replacement ratio, randomly selected words in a portion of the high-resource (English) training corpus are swapped with their word-level translations, requiring no additional model training and only a bilingual vocabulary, which can be obtained at near-zero cost for virtually any language.
Evaluation on eight languages across five model sizes shows notable improvements on downstream tasks in the target language, with up to a 2x speedup in training to reach equivalent performance.
}

\metadata[Correspondence]{\sffamily Anastasiia Sedova: \url{asedova@apple.com}
}
\date{\sffamily\today}

\begin{document}

\maketitle

\section{Introduction}

Large language models (LLMs) have demonstrated remarkable performance across a wide range of language understanding and knowledge-intensive tasks~\citep{brown2020language,bubeck2023sparks,liu2024deepseek,team2023gemini}.
These results are enabled by pretraining on massive text corpora, sometimes comprising tens of trillions of tokens. 
At this scale, high-quality training data is realistically available only in English~\citep{li2024datacomp,penedo2024fineweb}, and LLM research has therefore largely concentrated on it.
Most other languages, in contrast, lack sufficient high-quality data in public web crawls: even for languages typically considered high-resource, the available data amounts to only 1–10\% of the data available for English, while for truly low-resource languages, it can fall below 1B tokens~\citep{penedo2025fineweb2pipelinescale}.
Since data should scale roughly linearly with the size of the model \citep{hoffmann2022training}, this makes it difficult to adapt and effectively train LLMs for such languages \citep{Pakray_Gelbukh_Bandyopadhyay_2025,veitsman-hartmann-2025-recent}.

One way to address this scarcity is to leverage data across other languages for multilingual modeling.
\citet{seto2025trainingbilinguallmsdata} demonstrate that the increased capacity of larger models consistently yields better performance than smaller language-specific models.
Such models also naturally map semantically similar concepts across languages into a shared representation space, enabling effective cross-lingual knowledge transfer~\citep{conneau_emerging_2020,hu-etal-2021-explicit,liu-niehues-2025-middle}. %
One of the factors impacting this alignment is the presence of pretraining samples in which similar concepts from different languages co-occur in the same sentence, forcing the model to predict tokens in one language from tokens in another~\citep{wang-etal-2025-investigating-scaling}.
Existing methods that increase such cross-lingual mixing typically rely on translation systems, larger teacher models, or substantial parallel corpora \citep{wang-etal-2025-investigating-scaling,yoo-etal-2025-code-switching,li-etal-2024-prealign}, which are often unavailable for the data-constrained languages that would potentially benefit most.

\begin{figure}
    \centering
    \includegraphics[width=\linewidth]{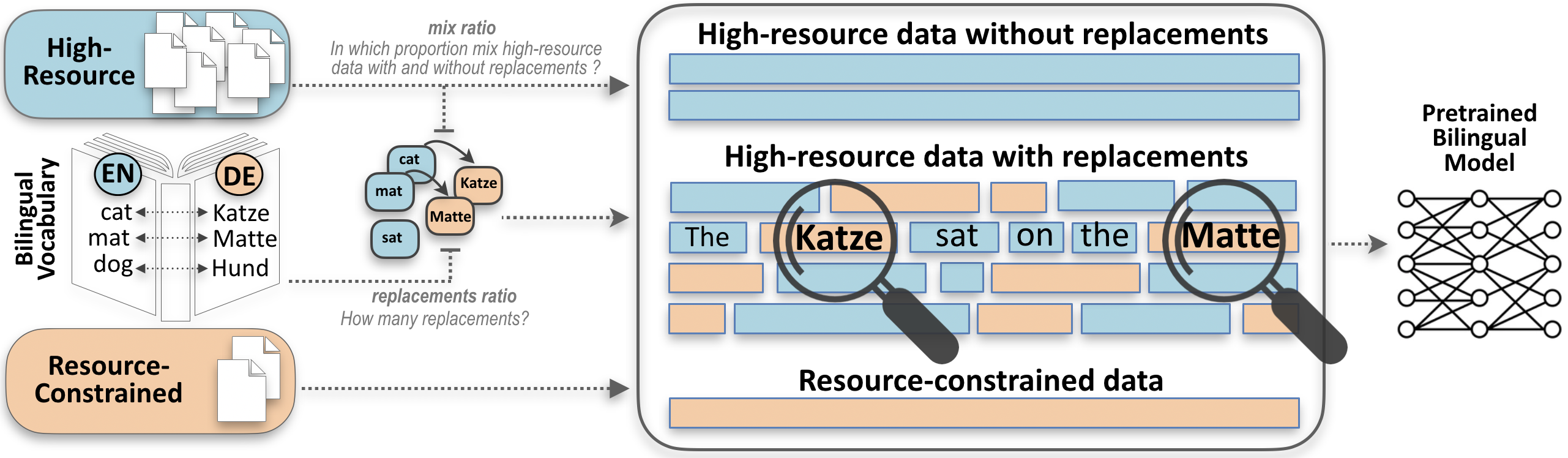}
    \caption{Overview of \texttt{LINK}. Using a bilingual vocabulary, we substitute randomly selected words in a portion of the high-resource pretraining data with their translations in the low-resource language. The \textit{mix ratio} controls what fraction of the high-resource data is replaced, and the \textit{replacement ratio} determines the amount of replacements.}
    \label{fig:overview}
\end{figure}

We introduce \textbf{\texttt{LINK}} (\textbf{L}exical \textbf{IN}terventions for \textbf{K}nowledge transfer): a method that facilitates cross-lingual knowledge transfer\footnote{In context of this paper, following previous work \citep{longpre2026atlas,li-etal-2024-prealign,10.5555/3454287.3454921}, by \textit{cross-lingual knowledge transfer}, we refer to the ability of a model pretrained on data from multiple languages to leverage knowledge (such as factual information or domain-specific concepts) acquired during pretraining from the data in one language when performing tasks in another.} through simple lexical substitutions in a multilingual pretraining mix (Figure \ref{fig:overview}). %
In contrast to previous work \citep{wang-etal-2025-investigating-scaling,yoo-etal-2025-code-switching,li-etal-2024-prealign}, our method requires only a small bilingual vocabulary to replace the words in a portion of the high-resource training data (controlled by the \textit{mix ratio}) with their low-resource translations up to a predefined proportion (\textit{replacement ratio}).
This makes \texttt{LINK} broadly applicable to virtually any language for which a bilingual vocabulary can be obtained (which, in practice, includes all written languages, even truly low-resource ones) and easy to integrate into pretraining pipelines at scale.

The experiments are conducted across five model sizes on four data-constrained settings simulated from high- and mid-resource languages by reducing the amount of training data. 
This setup allows greater flexibility to examine a broader range of scenarios and compare them against each other on reliable benchmarks, which are largely unavailable for truly low-resource languages.
We still validate our findings on four truly low-resource languages.
We also analyze the effect of bilingual vocabulary size on transfer performance and experiment with different placements of the interventions: \texttt{\textbf{LINK\_{uni}}} makes replacements on a randomly selected part of the data, and \texttt{\textbf{LINK\_{domain}}} intervenes only on the domain-specific part, thus enabling targeted domain knowledge transfer while preserving most of the high-resource data.
Our experiments demonstrate that such targeted interventions enable the model to maintain high performance in both languages.
An ablation study further shows that \texttt{LINK} facilitates cross-lingual knowledge transfer even when interventions are applied to data unrelated to the target domain.

Overall, using \texttt{LINK}, we show:  

\begin{itemize}%
    \item simple word-level substitutions 
    improve cross-lingual knowledge transfer from high-resource to data-constrained language during pretraining, yielding better downstream performance in data-constrained language and up to $2\times$ speedup in training under imbalanced data settings; %
    \item targeted interventions on a domain-specific portion of the high-resource training data lead to improvements comparable to intervening on the entire corpus, while maintaining high downstream performance in both languages;
    \item interventions on data not directly related to the end task still contribute to cross-lingual knowledge transfer and improve downstream performance.
\end{itemize}

\section{Related Work}

\paragraph{Data Limitations for Multilingual Language Models}
Training language models requires data and model sizes to scale jointly \citep{hoffmann2022training}. As a result, the development of multilingual models has been closely tied to the availability of large-scale multilingual corpora. 
Widely used multilingual datasets \citep{wenzek2020ccnet,xue2021mt5,penedo2025fineweb2pipelinescale}
provide web-crawled text in over 100 languages, but with a dramatic long-tail distribution:
for most languages,
data volumes can be 10–100x smaller than English, making it difficult to train even moderately sized models.
This data scarcity propagates through to model performance: models such as BLOOM \citep{workshop2022bloom}, Llama~2 \citep{touvron2023llama}, and Qwen~2.5 \citep{qwen2025qwen25technicalreport} have data mixtures with over 90\% containing English. 
This is further compounded by the curse of multilinguality: with fixed model capacity, adding more languages degrades per-language performance
\citep{conneau2020unsupervised}, disproportionately harming low-resource languages \citep{chang2024multilinguality}.

\paragraph{Cross-Lingual Knowledge Transfer}
Prior work has shown that multilingual models naturally learn to map semantically similar concepts across languages into a shared representation space~\citep{conneau_emerging_2020,hu-etal-2021-explicit,liu-niehues-2025-middle}.
Among the factors driving this is the co-occurence of similar concepts from different languages in the same context during training,
which helps the model learn to predict tokens in one language from tokens in another 
\citep{luong-etal-2015-bilingual,wang-etal-2025-investigating-scaling}, and 
even lightweight cross-lingual signal can meaningfully improve transfer \citep{li-etal-2024-prealign}.
Such multilingual knowledge transfer is particularly important when target language data is limited:
\citet{tanzer-etal-2024-benchmark} provide a benchmark for learning to translate a new language from a single book, %
and \citet{seto2025trainingbilinguallmsdata} demonstrated that bilingual models trained with limited target-language data benefit from scaling auxiliary high-resource data.
These and other works show that high-resource data can improve low-resource performance, but doing so without large parallel corpora or auxiliary models remains challenging.

\paragraph{Word- and Segment-Level Interventions}
Simple word-level perturbations have seen success as an effective data augmentation techniques in natural language processing. \citet{xie2017data} show that noisy word substitution improves language models and machine translation systems,
\citet{wei2019eda} further demonstrate that simple operations such as synonym replacement and random insertion improve text classification in low-data regimes.
In machine translation, bilingual dictionaries have been used to improve translation of rare words \citep{fadaee2017data}, while random word replacements have been shown to regularize neural machine translation models \citep{wang2018switchout}.
\citet{kobayashi2018contextual} extend word substitution by using a language model to predict contextually appropriate replacements. 
These works establish that even simple lexical perturbations provide meaningful training signal. 
Recent work with LLM-generated code-swtiched data included in pretraining further suggests that code-switching abilities may be key to improving multilingual capabilities \citep{wang-etal-2025-investigating-scaling, yoo-etal-2025-code-switching,li-etal-2024-prealign}.

Our \texttt{LINK} method is most closely related to these methods but differs in two key respects. First, we make no explicit code-switching assumption. \texttt{LINK} is motivated by the topology of today's LLMs: inspired by older research in multilingual word embeddings \citep{luong-etal-2015-bilingual}, we observe that by generating token representations using a bilingual context, we can enable multilingual representations of tokens for monolingual inputs.
Second, \texttt{LINK} is lightweight: it requires no parallel corpora, no auxiliary models, and no additional training stages, but only a bilingual vocabulary obtainable at near-zero cost, making it applicable to truly low-resource settings where none of these resources are available.

\section{Lexical Interventions for Cross-Lingual Knowledge Transfer}
\label{sec:method}

We assume a data-constrained\footnote{We use \textit{data-constrained} rather than \textit{low-resource}, but retain the standard LR abbreviation for brevity. See Appendix \ref{sec:app:dc_lr} for additional discussion.} target language dataset $D_{LR}$ of limited size and a high-resource language dataset $D_{HR}$ available in effectively unlimited quantity\footnote{Although \texttt{LINK} has no formal limitation on the number of languages, we follow \citet{seto2025trainingbilinguallmsdata} and focus on the bilingual setup for controllability
and leave other experiments for future work.}.
Before training, we apply lexical interventions to a portion of $D_{HR}$: %
using a bilingual vocabulary $V_{HR \leftrightarrow LR}$, we randomly replace words in $D_{HR}$ with their target-language translations. For a sample $x = (w_1, \ldots, w_n) \in D$, where $D \subseteq D_{HR}$ is the subset selected for intervention, let $k_x \in [0, n]$ denote the number of tokens to replace. The target \textit{replacement ratio} $r \in [0,1]$ controls what fraction of words in each sample are replaced. The intervened data is:
\begin{equation}
    D_{HR+LR} = \left\{ \textsc{Replace}(x, V_{HR \leftrightarrow LR}, k_x, r) \mid x \in D \right\},
\end{equation}
where $\textsc{Replace}(x, V, k, r)$ swaps $k$ randomly selected tokens in $x$ with their translations from $V$, leaving other words unchanged. Since bilingual vocabularies vary in coverage, the actual number of replacements may fall below $r$.

An example of an English sample before and after \texttt{LINK} for German-English mix, 70\% replacements: \\

\begin{mdframed}[backgroundcolor=gray!10]
  \begin{quote}    
\textit{
[\dots]
Combine the lamb with the onion mixture.
Add the cinnamon, oregano and red wine and cook for a few minutes.
Add the tomatoes and a cup of water or stock.
[\dots]
See more Greek recipes.
[\dots]
\newline
\newline
[\dots]\textbf{Kombinieren} \textbf{der} \textbf{Lamm} \textbf{mit} \textbf{der} \textbf{Zwiebel} mixture. 
Add \textbf{der} \textbf{Zimtbaum}, oregano \textbf{und} \textbf{rot} \textbf{Wein} \textbf{und} \textbf{Koch} \textbf{da} a \textbf{wenig} minutes. 
Add \textbf{der} tomatoes \textbf{und} a \textbf{Tasse} \textbf{aus} \textbf{Wasser} \textbf{oder} \textbf{Vorrat}.[\dots]\textbf{Sehen} \textbf{mehr} Greek recipes.[\dots]
}
\end{quote}
\end{mdframed}

Importantly, we aim to create cross-lingual co-occurrences rather than well-formed bilingual sentences; that is why we do not filter for grammatical correctness or translation accuracy. 
This approach keeps the intervention computationally cheap and turns it into a simple preprocessing step with negligible overhead requiring only a dictionary lookup per token.

We primarily consider two replacement strategies:
\begin{itemize}
    \item \textbf{Uniform interventions (\texttt{LINK\_{uni}})}, where 
    the words are replaced across a randomly selected portion of $D_{HR}$.
    From $D_{HR}$, we randomly select a subset $D_{HR}^{\text{rand}}$, which amount is defined by \textit{mix ratio}, and replace up to $r$ fraction of words in each sentence $x \in D$ with their translations from $V_{HR\leftrightarrow{}LR}$. 
    The remaining portion of the high-resource data, $D_{HR}^{\prime} = D_{HR} \setminus D_{HR+LR}^{\text{rand}}$, is included in the training set without modifications, as is the data-constrained $D_{LR}$, which are not subject to replacements to not reduce the (already highly limited) data.
    The resulting training dataset is defined as:
    \begin{equation}
        D_{\text{train}} = D_{LR} \cup D_{HR+LR}^{\text{rand}} \cup D_{HR}^{\prime}.
    \end{equation}
    \item 
    \textbf{Domain-specific interventions (\texttt{LINK\_{domain}})}, which applies interventions only to the \textit{task-} or \textit{domain-specific} portion of the high-resource data. 
    This is motivated by the practical observation that domain-specific knowledge is often abundant in high-resource language but scarce in the data-constrained languages (e.g., scientific content), making it a natural candidate for targeted transfer. 
    Under this more conservative setting, we address the lack of such domain data in the target language dataset, while we minimize the intervention cost on the high-resource dataset (which inevitably grows with increasing the size of intervened high-resource data).
    Let $D_{HR}^{\text{task}} \subset D_{HR}$ denote the domain-specific portion of the high-resource data, and $D_{HR}^{\text{non-task}}:=D_{HR}\setminus D_{HR}^{\text{task}}$.
    The domain-specific mixed subset $D_{HR+LR}^{\text{task}}$ is created by making replacements in $D_{HR}^{\text{task}}$ using $V_{HR\leftrightarrow{}LR}$. 
    The remaining data are included in the training set without modifications, resulting in:
    \begin{equation}
        D_{\text{train}} = D_{LR} \cup D_{HR+LR}^{\text{task}} \cup D_{HR}^{\text{non-task}}.
    \end{equation}
    This approach preserves the majority of the high-resource training data, concentrating cross-lingual signal where it is most needed while maintaining English performance.

\end{itemize}

\section{Experimental Setup}
\label{sec:exp}

English serves as the high-resource language for all experiments, as it is the only language with sufficiently large and diverse public web-scale datasets. The data-constrained scenario is simulated by subsampling four languages (German, French, Hindi, and Chinese) to approximately 350M tokens each (roughly 400K documents), comparable to prior work~\citep{seto2025trainingbilinguallmsdata} and closely mirroring the scale of genuinely low-resource languages in well-established multilingual datasets~\citep{penedo2025fineweb2pipelinescale,xue-etal-2021-mt5}. 
This enables comparison with having more target-language data, broader ablation studies, and evaluation on well-established benchmarks whose translation is often unfeasible for truly low-resource languages. 
Apart from that, we additionally experiment with four truly low-resource languages: Swahili, Yoruba, Amharic, and Igbo (see Section \ref{sec:true_LRL} for the results).
The English training data $D_{HR}$ is sampled from the FineWeb corpus~\citep{penedo2024fineweb} in quantities sufficient to avoid repetition within a single model's training.
All non-English data $D_{LR}$ is sampled from FineWeb2~\citep{penedo2025fineweb2pipelinescale}. 
A bilingual vocabulary for each target language $V_{HR\leftrightarrow{}LR}$ is constructed by extracting word entries and their corresponding English translations from a bilingual resource. 
While any bilingual dictionary or lexicon can serve this purpose, we use Wiktionary \citep{wiktionary} as it provides translation pairs for over 4,400 languages, 
making bilingual vocabularies readily available even for languages without parallel corpora or translation systems.
Dataset and vocabulary sizes are reported in Table~\ref{tab:data_stat}.

\begin{wraptable}{r}{5cm}                           
\centering                      
\small     
\vspace{-15pt} 
\setlength{\tabcolsep}{1pt}
\begin{tabular}{l@{\hskip 0.5pt}cc}
Language & \shortstack[c]{\footnotesize Bilingual\\[-4pt]\footnotesize vocabulary} & \shortstack[c]{\footnotesize Training\\[-4pt]\footnotesize Tokens} \\
\midrule
\midrule
English & -- & $\infty$\\
\midrule
German (DE) & 48,195 & 345M \\
Chinese (ZH) & 45,571 & 330M \\
French (FR) & 36,492 & 340M \\
Hindi (HI) & 25,001 & 342M \\
\hline
Swahili (SW) & 4,197 & 672M\\
Amharic (AM) & 1,487 & 435M \\
Yoruba (YO) & 675 & 96M \\
Igbo (IG) & 233 & 146M \\
\bottomrule
\end{tabular}
\vspace{-5pt} 
\caption{Bilingual vocabulary sizes and training data amounts.}
\vspace{-15pt}
\label{tab:data_stat}
  \end{wraptable}

We evaluate \texttt{LINK\_{uni}} and \texttt{LINK\_{domain}} on zero-shot QA tasks: ARC Easy and Challenge \citep{clark2018think}, Hellaswag \citep{zellers2019hellaswag}, Lambada \citep{paperno2016lambada}, PiQA \citep{bisk2020piqa}, SciQ \citep{welbl2017crowdsourcing}, and Winogrande \citep{sakaguchi2021winogrande}.  These are knowledge-based tasks that small models with limited data still perform well on.
Non-English evaluations are conducted via translation of the original dataset.  
We translate primarily using our own systems to ensure that translation artifacts will be consistent
throughout. %

We train decoder-only GPT models at five scales (137M, 345M, 760M, 1.3B, and 2.7B parameters) using the Megatron-LM framework with the Aya multilingual tokenizer  (250K vocabulary)~\citep{ustun2024aya} and a sequence length of 1024. 
Training runs for 10K, 30K, 50K, and 100K steps, correspondingly, with a batch size of 1024 samples. 
In a separate set of experiments, we empirically determined an optimal ratio for combining English and LR language data of 97.5\% (high-resource) to 2.5\% (data-constrained)\footnote{The data mix ratio of 97.5:2.5 was selected from a range of 50:50 to 99:1 based on final checkpoint perplexity in preliminary English–German experiments}.

\section{Results}
We compare \texttt{LINK} against three baselines:
(1) \textit{LR (UB)} as an upper bound, trained with enough target language data for one training epoch, serving as an oracle and illustrating the gap that data scarcity creates\footnote{Not available for Hindi due to limited data available in the Fineweb2 dataset.};
(2) \textit{LR} as a lower bound, trained solely on the scarce amount of resource-constrained data; and 
(3) \textit{LR + HR} trained on a mixture of the scarce quantity of resource-constrained data and sufficient quantity of high-resource data.
Due to space constraints, we restrict our discussion to results with the 1.3B parameter models, and defer results from smaller (137M, 345M, and 760M parameters) and larger (2.7B parameters) models to the Appendix \ref{sec:app:models}.

\subsection{Uniform Interventions}
\label{sec:uniform_intervention}
Table~\ref{tab:main_1_3B} reports evaluation results of our \texttt{LINK\_{uni}} experiments.
All results are presented for the mix ratio of 90 and replacement ratio of 70 based on the ablation study. 
In practice, however, vocabulary coverage often caps the actual per-sample replacement rate below the 70\% target, meaning \textit{every} word that can be replaced is replaced under this setup (see more details in Appendix \ref{sec:app:target_actual_repl}).
The results of other replacement configurations are discussed in Section \ref{sec:ablation}.

\begin{table*}[t!]
\centering
\small
\setlength{\tabcolsep}{2.7pt}                               
\begin{tabular}{l|ccccccc!{\vrule width 1pt}c||ccccccc!{\vrule width 1pt}c}
&
\multicolumn{8}{c||}{\textbf{LR performance}} &
\multicolumn{8}{c}{\textbf{HR performance}} \\
\cline{1-17}

\textbf{Setup} &
\textbf{A-E} &
\textbf{A-C} &
\textbf{HS} &
\textbf{LB} &
\textbf{PQ} &
\textbf{SQ} &
\textbf{WG} &
\textbf{Avg} &
\textbf{A-E} &
\textbf{A-C} &
\textbf{HS} &
\textbf{LB} &
\textbf{PQ} &
\textbf{SQ} &
\textbf{WG} &
\textbf{Avg}
\\
\hline
\multicolumn{17}{c}{\cellcolor{langDElight}\textbf{German}} \\
\hline
\rowcolor{langDEvlight}
LR (UB) &
37.2 & 28.6 & 41.9 & 28.7 & 62.8 & 65.9 & 51.9 & 45.3 & 33.5 & 24.1 & 33.5 & 28.9 & 60.2 & 61.3 & 52.1 & 41.9 \\
\hdashline
\rowcolor{langDElight}
LR &
30.8 & 22.7 & 29.1 & 12.7 & 55.2 & 47.3 & 52.0 & 35.7 & 27.8 & 23.5 & 26.8 & 4.5 & 52.2 & 43.1 & 52.3 & 32.9 \\
\rowcolor{langDElight}
LR + HR &
38.0 & 25.8 & 37.9 & 26.7 & 59.0 & 64.1 & 52.7 & 43.5 & 52.1 & 27.8 & \textbf{57.6} & 53.0 & \textbf{74.4} & 74.6 & 53.7 & 56.2 \\
\hline
\rowcolor{langDE}
{\texttt{\textbf{LINK\_{uni}}}} &
\textbf{40.3} & \textbf{26.6} & \textbf{39.4} & \textbf{26.9} & 59.8 & 64.7 & \textbf{53.5} & \textbf{44.5} & 50.0 & 27.4 & 54.3 & 49.4 & 71.4 & 72.3 & 56.4 & 54.5 \\
\rowcolor{langDE}
{\texttt{\textbf{LINK\_{domain}}}} &
39.3 & 25.0 & 38.1 & 26.6 & \textbf{60.3} & \textbf{66.7} & 52.0 & 44.0 & \textbf{52.6} & \textbf{29.2} & 57.0 & \textbf{53.4} & 73.6 & \textbf{76.6} & \textbf{56.7} & \textbf{57.0} \\
\hline
\multicolumn{17}{c}{\cellcolor{langFRlight}\textbf{French}} \\
\hline
\rowcolor{langFRvlight}
LR (UB) &
41.3 & 27.4 & 46.7 & 32.5 & 66.0 & 66.3 & 54.9 & 47.9 & 38.1 & 21.8 & 33.6 & 30.6 & 62.0 & 66.5 & 53.2 & 43.7 \\
\hdashline
\rowcolor{langFRlight}
LR &
31.9 & 21.2 & 30.1 & 10.5 & 55.7 & 54.4 & 53.2 & 36.7 & 28.2 & 21.7 & 26.6 & 4.3 & 52.0 & 46.6 & 49.9 & 32.8 \\
\rowcolor{langFRlight}
LR + HR &
38.4 & 24.9 & 40.2 & 28.3 & 60.7 & \textbf{65.1} & 54.2 & 44.5 & \textbf{53.3} & 28.5 & \textbf{57.8} & 52.4 & \textbf{74.7} & \textbf{77.4} & 54.9 & \textbf{57.0} \\
\hline
\rowcolor{langFR}
{\texttt{\textbf{LINK\_{uni}}}} &
39.8 & 25.8 & \textbf{42.2} & \textbf{30.4} & 60.6 & 63.4 & \textbf{54.2} & \textbf{45.2} & 48.7 & 27.1 & 53.8 & 49.5 & 71.6 & 70.9 & 53.9 & 53.6 \\
\rowcolor{langFR}
{\texttt{\textbf{LINK\_{domain}}}} &
\textbf{40.4} & \textbf{26.2} & 41.3 & 29.2 & \textbf{61.0} & 64.2 & 52.9 & 45.0 & 52.0 & \textbf{29.2} & 57.3 & \textbf{53.0} & 73.5 & 75.3 & \textbf{55.0} & 56.5 \\
\hline
\multicolumn{17}{c}{\cellcolor{langZHlight}\textbf{Chinese}} \\
\hline
\rowcolor{langZHvlight}
LR (UB) &
46.1 & 26.6 & 39.0 & -- & 59.6 & 78.6 & 51.8 & 50.3 & 34.4 & 19.9 & 29.0 & 15.4 & 55.5 & 64.3 & 52.7 & 38.8 \\
\hdashline
\rowcolor{langZHlight}
LR &
33.5 & 23.7 & 30.5 & -- & 53.8 & 62.9 & 49.4 & 42.3 & 27.3 & 20.3 & 26.7 & 1.4 & 50.8 & 40.1 & 48.7 & 30.7 \\
\rowcolor{langZHlight}
LR + HR &
39.7 & 24.6 & 37.3 & -- & 58.3 & 75.5 & \textbf{51.5} & 47.8 & \textbf{53.2} & \textbf{29.7} & \textbf{56.4} & 50.2 & \textbf{73.8} & 73.8 & 55.2 & \textbf{56.0} \\
\hline
\rowcolor{langZH}
{\texttt{\textbf{LINK\_{uni}}}} &
\textbf{41.8} & \textbf{26.8} & \textbf{38.3} & -- & \textbf{59.5} & 76.4 & 51.2 & \textbf{49.0} & 49.7 & 27.1 & 54.3 & 50.0 & 73.0 & 73.0 & \textbf{56.3} & 54.7 \\
\rowcolor{langZH}
{\texttt{\textbf{LINK\_{domain}}}} &
41.5 & 25.4 & 37.8 & -- & 59.0 & \textbf{77.3} & 51.0 & 48.7 & 50.3 & 27.1 & 56.2 & \textbf{51.9} & 73.8 & \textbf{74.5} & 55.2 & 55.6 \\
\hline
\multicolumn{17}{c}{\cellcolor{langHIlight}\textbf{Hindi}} \\
\hline
\rowcolor{langHIlight}
LR &
30.6 & 22.9 & 28.3 & -- & \textbf{55.0} & -- & \textbf{52.9} & 37.9 & 27.3 & 22.4 & 25.9 & 2.0 & 51.0 & 36.4 & 51.0 & 30.9 \\
\rowcolor{langHIlight}
LR + HR &
33.3 & 24.1 & 30.4 & -- & 53.0 & -- & 51.0 & 38.4 & \textbf{53.4} & \textbf{28.2} & \textbf{57.9} & \textbf{52.7} & \textbf{74.0} & \textbf{77.9} & \textbf{56.3} & \textbf{57.2} \\
\hline
\rowcolor{langHI}
{\texttt{\textbf{LINK\_{uni}}}} &
35.0 & 24.6 & \textbf{31.8} & -- & 54.0 & -- & 52.5 & \textbf{39.6} & 48.1 & 27.3 & 52.5 & 47.5 & 72.1 & 75.3 & 53.2 & 53.7 \\
\rowcolor{langHI}
{\texttt{\textbf{LINK\_{domain}}}} &
\textbf{35.5} & \textbf{25.3} & 30.9 & -- & 49.0 & -- & 50.3 & 38.2 & 52.4 & 27.4 & 56.4 & 52.5 & 74.0 & 76.6 & 55.8 & 56.4 \\
\hline
\end{tabular}
\caption{{\texttt{\textbf{LINK\_{uni}}}} and {\texttt{\textbf{LINK\_{domain}}}} 1.3B model results across German, French, Chinese, and Hindi. Bold indicates best per task-language pair among LR~(S), LR+HR, {\texttt{LINK\_{uni}}}, and {\texttt{LINK\_{domain}}} (excluding the LR~(UB)).}
\label{tab:main_1_3B}
\end{table*}

\footnotetext{Note that perplexity values are not directly comparable across languages due to differences in tokenizer fertility. Full scores are provided in Appendix \ref{sec:app:ppl}.}

\texttt{LINK\_{uni}} yield consistent improvements in data-constrained language performance across all four languages, outperforming both LR and LR+HR baselines, often by a substantial margin.
Remarkably, using only a fraction of the data-constrained data available to LR (UB), \texttt{LINK\_{uni}} even surpasses this upper bound baseline (e.g.,~ARC-Easy for German). 
This proves our hypothesis that simple word-level replacements can unlock cross-lingual knowledge transfer between languages powerful enough to close the gap with models trained on significantly more target data. 

\begin{wrapfigure}{rt!}{0.4\textwidth}
\vspace{-15pt}
\centering
\includegraphics[width=6cm]{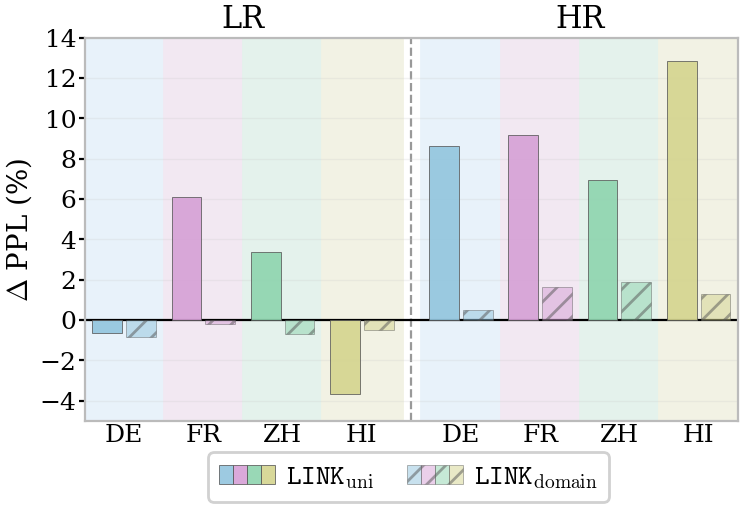}
\caption{{\texttt{LINK\_{uni}}} and {\texttt{LINK\_{domain}}} percentage change in perplexity relative to the LR+HR baseline for 1.3B models, computed as $\frac{\text{PPL}_{\texttt{LINK\_{uni}}} - \text{PPL}_{\text{LR+HR}}}{\text{PPL}_{\text{LR+HR}}}$\protect\footnotemark.}
\label{fig:ppl}
\vspace{-30pt}
\end{wrapfigure}
However, these gains come at the cost of high-resource language performance: English downstream scores drop by up to 5.5pp compared to the LR+HR baseline. %
Figure~\ref{fig:ppl} also confirms this: massive uniform interventions performed by \texttt{LINK\_{uni}} lead to a substantial increase in English perplexity across all bilingual setups, reflecting the degradation caused by modifying a large portion of the high-resource training data.
The effect on data-constrained language perplexity is less consistent — it decreases for some languages but increases for others, suggesting that downstream gains from cross-lingual knowledge transfer do not necessarily correlate with lower general-domain perplexity.

\begin{figure}
\centering
\includegraphics[width=15cm]{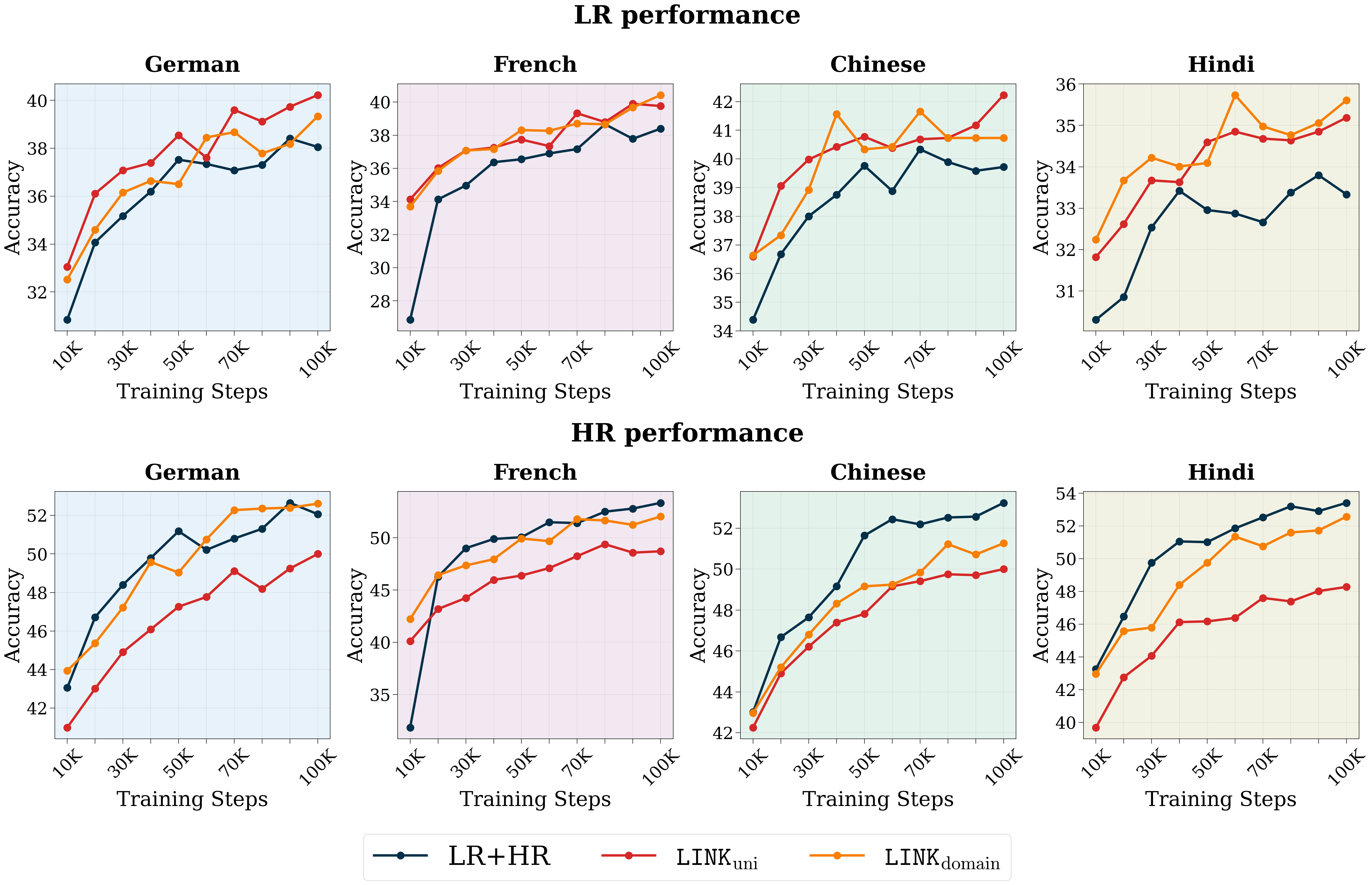}
\caption{{\texttt{\textbf{LINK\_{domain}}}}: ARC-Easy accuracy for 1.3B models. Top: target language evaluation; bottom: English evaluation.
}
\label{fig:arc_easy_domain}
\end{figure}

\subsection{Domain-Specific Interventions}
\label{sec:domain_intervention}

Next, we experiment with \texttt{LINK\_{domain}}, where interventions are applied only to domain-specific data.
In contrast to the \texttt{LINK\_{uni}} setup, the only relevant data for replacements in \texttt{LINK\_{domain}} is the domain-specific subset, and, within it, the \textit{mix ratio} equals 100, while the remaining English data is left unmodified. Our primary motivation for including a domain specific intervention is to limit the amount of data %
that is replaced, to protect performance in English.
As a representative domain, we focus on scientific content, targeting the ARC downstream task~\citep{clark2018think}. 
To identify the relevant training data, following \citet{grangier2025taskadaptive,seto2025trainingbilinguallmsdata} we cluster the English data using k-means with multilingual BERT representations~\citep{devlin-etal-2019-bert} into 32 clusters.
The resulting clusters show clear topic-wise separation, with one cluster (comprising roughly 5.3B tokens) corresponding to scientific knowledge -- confirmed by assigning English and German ARC validation sets to the same centroids, which are almost entirely attributed to this cluster (see Appendix~\ref{sec:app:clustering}). 
Therefore the interventions were applied to all samples attributed to this cluster while the other samples remains intact.  
We preserved the original cluster distribution in the pretraining mix.

The results are provided in Table~\ref{tab:main_1_3B}.
Additionally, Figure~\ref{fig:arc_easy_domain} shows ARC-Easy (i.e., target domain) accuracy throughout training for all four languages. %
\texttt{LINK\_{domain}} interventions match or exceed \texttt{LINK\_{uni}} interventions on target-language performance (top row) while preserving English performance (bottom row), unlike \texttt{LINK\_{uni}}, which causes a notable drop.
Importantly, with both intervention strategies, \texttt{LINK} improves training efficiency for the target language. 
For example, we see that for German, the \texttt{LINK} models reaches the baseline's final performance at ~40k steps versus ~80k, representing an approximately $2\times$ speedup, while for Hindi both intervention strategies surpass the baseline as early as 30k steps.
Furthermore, Figure~\ref{fig:ppl} shows that \texttt{LINK\_{domain}} result in negligible perplexity changes for both languages, in stark contrast to the substantial increases in English perplexity observed with \texttt{LINK\_{uni}} interventions. 
Together, these results show that targeting interventions to domain-relevant data achieves strong data-constrained language transfer without sacrificing high-resource performance.

\subsection{Low-Resource Experiments}
\label{sec:true_LRL}
Evaluating on truly low-resource languages introduces additional challenges: severely limited data limits bilingual vocabulary size and training corpus (e.g., Yoruba has only 96M tokens in FineWeb2 and 1.4\% of the German vocabulary, see Table \ref{tab:data_stat}) constraining both what the model can learn from the target language directly, and the degree of overlap the two languages have in the data.
Apart from that, standardized evaluation benchmarks are largely unavailable, as most established English benchmarks lack translations for these languages and automatic translation is often unreliable due to the same data 
scarcity being not enough to train reliable machine translation systems. 
Evaluation is therefore restricted to the few benchmarks with existing multilingual versions.

\begin{wraptable}{l}{6.5cm}
\centering
\small
\begin{tabular}{p{0.06cm}|l|p{0.5cm}|p{0.5cm}|p{0.5cm}|p{0.5cm}}
& & {\textbf{AM}} & {\textbf{IG}} & {\textbf{SW}} & {\textbf{YO}} \\
\cline{1-6}
\multirow{4}{*}{\rotatebox{90}{1.3B}} & LR & \textbf{44.7} & 60.1 & 47.8 & 38.2 \\
& LR + HR & 42.1 & \textbf{61.6} & 48.5 & 40.0 \\
\cline{2-6}
& {\texttt{\textbf{LINK\_{uni}}}} & 41.6 & 60.4 & 48.5 & \textbf{40.7} \\
& {\texttt{\textbf{LINK\_{domain}}}} & 40.9 & 60.4 & \textbf{50.1} & 38.7 \\
\cline{1-6}
\multirow{4}{*}{\rotatebox{90}{345M}} & LR & 42.5 & 51.5 & 46.5 & 38.1 \\
& LR + HR & 41.3 & 57.5 & 46.4 & 34.8 \\
\cline{2-6}
& {\texttt{\textbf{LINK\_{uni}}}} & \textbf{42.4} & \textbf{59.1} & 46.8 & \textbf{39.7} \\
& {\texttt{\textbf{LINK\_{domain}}}} & 41.7 & 58.1 & \textbf{47.4} & 33.2 \\
\cline{1-6}
\end{tabular}
\caption{\texttt{LINK} average target-language benchmark results for true low-resource languages.}
\label{tab:true_lrl_avg_2}
\end{wraptable}

Table \ref{tab:true_lrl_avg_2} reports the average \texttt{LINK} results for four low-resource languages: Amharic, Igbo, Swahili, and Yoruba.
The evaluation is done on the benchmarks that are available for these languages (see the exact list and full scores in Appendix~\ref{sec:app:true_LRL}).
The effect of domain interventions varies across languages.  
For Igbo, Swahili, and Yoruba, we see improvement in all setting except 1.3B Igbo setup.
For Swahili, domain-specific interventions yield consistent improvements at both scales, while for Yoruba and Igbo, uniform interventions perform best. This can be explained by much larger vocabulary available for Swahili (4197) than for other languages, which motivates our next study of bilingual vocabulary size (see the next section).
In the Amharic, the LR only baseline performs the best consistently.  
This may reflect the greater script and typological distance of Amharic from English, which limits the effectiveness of lexical replacement and, more generally, training with English data.

\section{Ablation Studies}
\label{sec:ablation}

\texttt{LINK} introduces several design choices including the bilingual vocabulary, replacement ratio, mix ratio, and domain relevance. To understand which of these factors most strongly influence target language performance and to guide practitioners in applying our method, we conduct a series of ablation studies.

\paragraph{Model Sizes}                                
We evaluate across five model sizes (137M--2.7B parameters) to verify that our method's benefits are not limited to a single scale.
Both \texttt{LINK\_{uni}} and \texttt{LINK\_{domain}} consistently match or outperform the LR+HR baseline at every scale.
Benefits emerge even at 137M and become increasingly pronounced at larger scales: the gap between \texttt{LINK} and the baseline widens as model size grows, with the largest improvements observed at 760M and 1.3B.
This trend is consistent across both close (German, French) and distant (Chinese, Hindi) language pairs, suggesting that larger models are better able to exploit the cross-lingual signal introduced by our method.
Full results are reported in Appendix~\ref{sec:app:models}. 

\paragraph{Reduced Vocabulary Size}
\begin{wrapfigure}{r}{0.5\textwidth}
\vspace{-10pt}
\centering
\includegraphics[width=8cm]{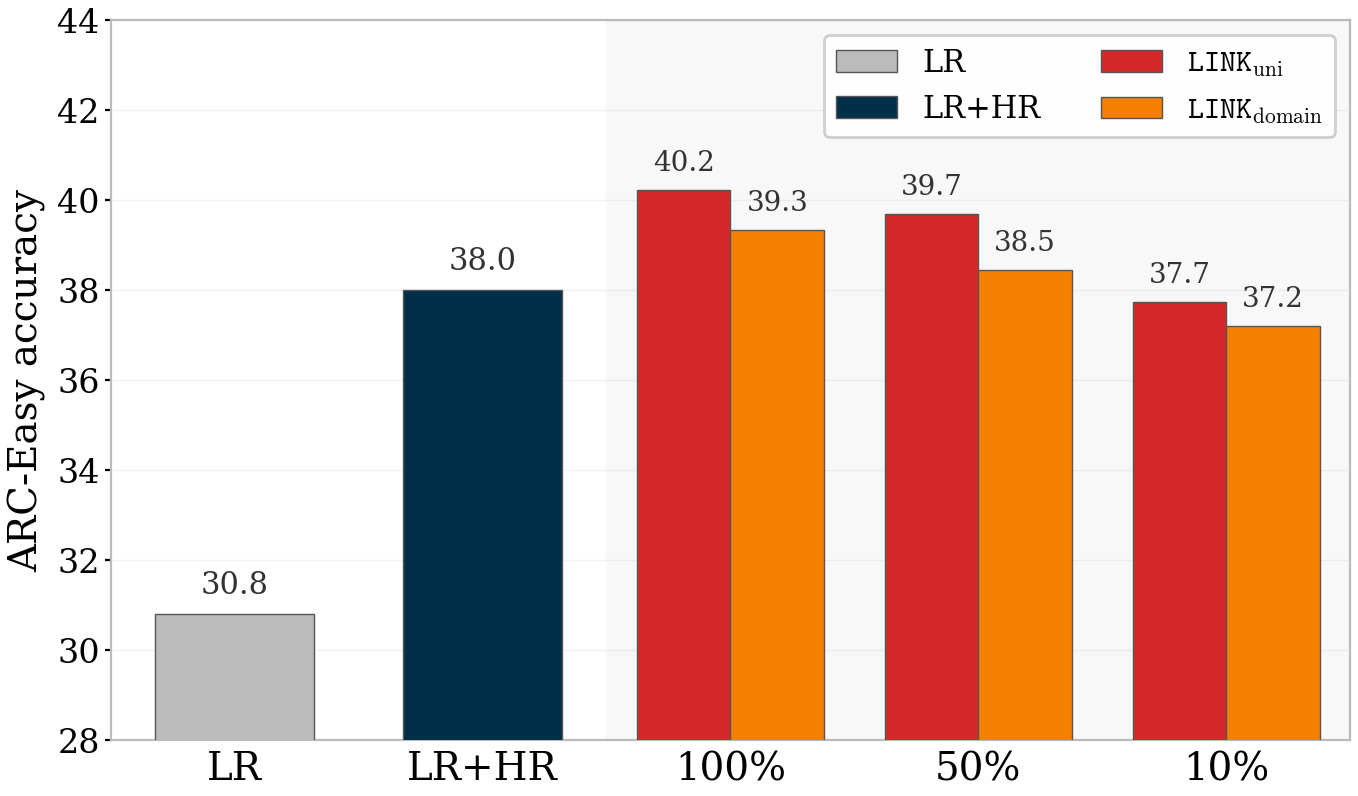}
\caption{\texttt{LINK} ARC-Easy accuracy on German with reduced bilingual vocabularies (1.3B model).}
\label{fig:reduced_vocab}
\end{wrapfigure}
To examine the effect of vocabulary size on transfer performance, we conduct experiments with reduced German bilingual vocabularies to 50\% and 10\% of the original size (each vocabulary is a subset of a larger one). 
Figure~\ref{fig:reduced_vocab} presents the results on ARC-Easy for both uniform and domain-specific settings. 
Both settings show a similar pattern: full vocabulary (100\%)  outperforms the LR+HR baseline, while reducing the vocabulary to 50\% (around 24,000 word pairs) leads to a modest decline, which is amplified by further reduction to~10\% (around 4,800~word pairs). 
These results indicate that vocabulary size is an important factor in transfer performance, with larger bilingual dictionaries yielding stronger gains.  
This corroborates inconsistent gains we observe for some of the low-resource languages as the vocabulary size is 9\% and lower of the German vocabulary for the four low resource languages we show.
Additionally, we reduced the vocabulary even further to 1\% - results of these experiments are provided in Appendix \ref{sec:app:vocab_reduction}.
Encouragingly, obtaining vocabularies of this size is not a significant barrier, as Wiktionary alone provides at least 1,000 translation pairs for 186 languages and over 10,000 for 69.

\paragraph{Replacement \& Mix Ratio}

\renewcommand{\arraystretch}{0.9}
\begin{wraptable}{r}{0.32\textwidth}
\centering
\small 
\setlength{\tabcolsep}{1pt}
\begin{tabular}{cc|c|cc}
\textbf{Repl} & \textbf{Mix} & \textbf{PPL\textsubscript{EN}} & \textbf{PPL\textsubscript{DE}} & \textbf{A-E\textsubscript{DE}} \\
\hline
\multicolumn{2}{c|}{LR} & 265.4 & 58.1 & 30.8 \\
\multicolumn{2}{c|}{LR + HR} & 10.0 & 16.6 & 38.1 \\
\hline
10 & 10 & 10.0 & 16.6 & 38.5 \\
30 & 30 & 10.1 & 16.6 & 38.9 \\
50 & 70 & 10.4 & 16.5 & 38.5 \\
50 & 90 & 10.6 & 16.5 & 39.6 \\
70 & 70 & 10.5 & 16.6 & 38.6 \\
70 & 90 & 10.9 & \textbf{16.5} & \textbf{40.2} \\
\end{tabular}
\caption{\texttt{LINK} across varying replacement (\textbf{Repl}) and mixing (\textbf{Mix}) ratios (English--German, 1.3B).}
\vspace{-20pt}
\label{tab:repl_mix_ablation}
\end{wraptable}

Table~\ref{tab:repl_mix_ablation} presents an ablation over the replacement ratio and mixing ratio for the English–German pair, reporting English and German perplexity on FineWeb and FineWeb-2 validation sets, as well as ARC Easy accuracy in German. 
Low values for both ratios (10/10, 30/30) have minimal effect on downstream performance. 
Increasing either ratio improves results, though the mixing ratio appears to have a slightly stronger effect; the best configuration (70/90) achieves over 2 points above the non-augmented baseline. 
Notably, target-language perplexity remains stable across all configurations, while English perplexity increases by almost 1pp, further motivating \texttt{LINK\_{domain}}.

\label{sec:mix_repl_ratio}

\renewcommand{\arraystretch}{1}

\begin{wraptable}{l}{0.37\textwidth}
\vspace{-10pt}
\centering
\small
\setlength{\tabcolsep}{2pt}
\begin{tabular}{p{0.45cm}|l|c|cc}
& \textbf{Setup} & \textbf{PPL\textsubscript{EN}} & \textbf{PPL\textsubscript{LR}} & \textbf{A-E\textsubscript{LR}} \\
\hline
\multirow{4}{*}{DE}
& LR + HR     & 10.0 & 16.6 & 38.1 \\
\cline{2-5}
& Uniform     & 16.9 & 16.8 & 40.3 \\
& Domain      & \textbf{10.0} & \textbf{16.5} & 39.3 \\
& Non-domain  & 11.9 & 16.7 & \textbf{40.5} \\
\hline
\multirow{4}{*}{FR}
& LR + HR     & 9.9 & 8.9 & 38.4 \\
\cline{2-5}
& Uniform     & 10.8 & 9.4 & 39.8 \\
& Domain      & \textbf{10.1} & \textbf{8.9} & \textbf{40.4} \\
& Non-domain  & 11.8 & 9.4 & 39.8 \\
\hline
\multirow{4}{*}{ZH}
& LR + HR     & 10.0 & 42.4 & 39.7 \\
\cline{2-5}
& Uniform     & 10.7 & 43.9 & \textbf{41.8} \\
& Domain      & \textbf{10.2} & \textbf{42.1} & 41.5 \\
& Non-domain  & 12.3 & 45.4 & 40.8 \\
\hline
\multirow{4}{*}{HI}
& LR + HR     & 9.9 & 6.0 & 33.3 \\
\cline{2-5}
& Uniform     & 11.2 & \textbf{5.8} & 35.0 \\
& Domain      & \textbf{10.1} & 6.0 & 35.5 \\
& Non-domain  & 12.4 & 5.8 & \textbf{35.9} \\
\hline
\end{tabular}
\caption{Comparison of \texttt{LINK} intervention strategies (1.3B). PPL\textsubscript{EN}/PPL\textsubscript{LR}: perplexity on FineWeb/FineWeb-2 validation; A-E: ARC Easy
accuracy.}
\vspace{-7pt}
\label{tab:domain_vs_nondomain}
\end{wraptable}

\paragraph{Non-domain specific interventions and indirect knowledge transfer} 
To disentangle the effect of topical alignment between the replaced text and the target language from the effect of lexical exposure itself, we introduce a \textit{Non-domain} control experiment: interventions are applied to all data except for the domain-specific part.
That is, %
replacements are applied only to $D_{HR}^{\text{non-task}} = D_{HR} \setminus D_{HR}^{\text{task}}$, while the task-relevant data remains unchanged. The final training set is then $D_{\text{train}} = D_{LR} \cup D_{HR+LR}^{\text{non-task}}\cup D_{HR}^{\text{task}}$.
In our experimental setup from Section \ref{sec:domain_intervention}, the interventions on all non-scientific data means intervening on roughly the 95.5\% of the original English dataset -- i.e., more than domain-specific (4.5\%) or even uniform (90\%) interventions. 
As shown in Table~\ref{tab:domain_vs_nondomain}, this massive increase in intervention volume leads to further LR improvements for all four languages compared to the LR~+~HR baseline, with non-domain interventions outperforming both uniform and domain-specific setups for Hindi and German. 
This suggests that sheer volume of lexical exposure matters: the model benefits from encountering target-language vocabulary broadly across the training data, regardless of topical relevance. We observe an expected cost to English performance, but strikingly also observe that our approach unlocks domain transfer even without directly intervening on the relevant data.

\section{Conclusion}
\label{sec:conc}
This work proposes a data-level intervention method for improving language model pretraining in languages with scarce data. Our approach requires only a bilingual vocabulary, making it applicable at near-zero cost to over a thousand languages and at pretraining scale. Across different target languages and five model sizes, our method consistently improves downstream performance particularly for languages that are distant from the high-resource language, and remains effective even when interventions are applied to only a small, domain-specific
portion of the training data. These findings demonstrate a practical and scalable path toward building stronger language models for data-scarce languages.

\bibliographystyle{colm2026_conference}
\bibliography{colm2026_conference}

\appendix

\section{Appendix}

\subsection{Perplexity Results}
\label{sec:app:ppl}

Table \ref{tab:app:ppl} reports validation perplexity on FineWeb-2 (target language, LR) and FineWeb (English, EN) for each setup across all four model sizes.
At 137M, the smallest scale, \texttt{LINK\_{uni}} incurs a substantial English perplexity penalty (e.g., 36.49 vs.\ 29.53 for German), while \texttt{LINK\_{domain}} keeps English perplexity close to LR+HR across all languages.
Target-language perplexity remains comparable to the baseline for both strategies. 

Starting at 345M, \texttt{LINK\_{domain}} consistently achieves the lowest or near-lowest target-language perplexity while keeping English perplexity close to LR+HR. 
\texttt{LINK\_{uni}} matches or slightly exceeds the LR+HR English perplexity but shows competitive LR perplexity, with the best Hindi result.
This pattern strengthens at 760M, where \texttt{LINK\_{domain}} achieves the best LR perplexity for German and French, with English perplexity remaining within 0.1–0.2 points of LR+HR.
\texttt{LINK\_{uni}} also improves LR perplexity over the baseline (e.g., 19.09 vs.\ 20.37 for German) but at a larger English cost.
At 1.3B, the same trends hold: \texttt{LINK\_{domain}} closely matches or improves upon LR+HR on both sides, i.e., its English perplexity remains within 1 point of LR+HR, while its LR perplexity improves for German and French and matches LR+HR for Chinese. 
\texttt{LINK\_{uni}} yields comparable LR perplexity but incurs a slightly larger English penalty.
Across all scales, the LR-only baseline diverges for larger models, highlighting the instability of training on scarce data alone. 
         
\renewcommand{\arraystretch}{0.9}    

\begin{table*}[h!]                                                                                                  \centering
\small                                                                                                              \setlength{\tabcolsep}{5pt}                                                                                         \begin{tabular}{l|cc|cc|cc|cc}                                                                                      &                                                                                                                   \multicolumn{2}{c|}{\textbf{DE}} &                                                                                  \multicolumn{2}{c|}{\textbf{FR}} &                                                                                  \multicolumn{2}{c|}{\textbf{ZH}} &                                                                                  \multicolumn{2}{c}{\textbf{HI}} \\                                                                                  \cline{2-9}                                                                                                         \textbf{Setup} &                                                                                                    \textbf{LR} & \textbf{EN} &                                                                                         \textbf{LR} & \textbf{EN} &                                                                                         \textbf{LR} & \textbf{EN} &                                                                                         \textbf{LR} & \textbf{EN}                                                                                           \\                                                                                                                  \hline                                                                                                              \multicolumn{9}{c}{\cellcolor{gray!15}\textbf{137M}} \\                                                             \hline                                                                                                              LR (UB) & 24.54 & 86.90 & 13.38 & 94.03 & 54.62 & $>$10$^2$ & -- & -- \\
\hdashline                                                                                                          LR & \textbf{35.74} & $>$10$^2$ & \textbf{19.94} & $>$10$^2$ & $>$10$^2$ & $>$10$^2$ & \textbf{9.43} & $>$10$^2$ \\
LR + HR & 62.85 & \textbf{29.53} & 30.56 & \textbf{29.37} & $>$10$^2$ & \textbf{29.40} & 18.96 & \textbf{29.46} \\  \hline                                                                                                              \texttt{\textbf{LINK\_{uni}}} & 68.43 & 36.49 & 32.41 & 36.34 & $>$10$^2$ & 34.79 & 20.11 & 39.40 \\                \texttt{\textbf{LINK\_{domain}}} & 63.08 & 29.80 & 30.63 & 29.73 & $>$10$^2$ & 29.99 & 20.04 & 29.89 \\             \hline                                                                                                              \multicolumn{9}{c}{\cellcolor{gray!15}\textbf{345M}} \\                                                             \hline                                                                                                              LR (UB) & 10.02 & 32.15 & 6.61 & 32.95 & 14.63 & 67.43 & -- & -- \\                                                 \hdashline                                                                                                          LR & 29.40 & $>$10$^2$ & 19.94 & $>$10$^2$ & $>$10$^2$ & $>$10$^2$ & 7.87 & $>$10$^2$ \\                            LR + HR & 23.32 & 14.82 & 12.53 & 14.82 & 67.90 & 14.39 & 7.70 & 15.64 \\                                           \hline                                                                                                              \texttt{\textbf{LINK\_{uni}}} & 23.37 & 14.83 & 12.81 & 14.85 & 67.39 & 14.39 & \textbf{7.68} & 15.60 \\            \texttt{\textbf{LINK\_{domain}}} & \textbf{23.02} & \textbf{13.27} & \textbf{11.77} & \textbf{13.26} & \textbf{63.22} & \textbf{13.30} & 7.89 & \textbf{13.31} \\                                                  
\hline                                                                                                              \multicolumn{9}{c}{\cellcolor{gray!15}\textbf{760M}} \\                                                             \hline                                                                                                              LR (UB) & 8.39 & 25.55 & 5.75 & 25.94 & 11.09 & 51.25 & -- & -- \\                                                  \hdashline                                                                                                          LR & $>$10$^5$ & $>$10$^6$ & 28.45 & $>$10$^2$ & $>$10$^2$ & $>$10$^3$ & 9.66 & $>$10$^2$ \\                        LR + HR & 20.37 & \textbf{11.20} & 9.92 & \textbf{11.17} & 50.42 & \textbf{11.23} & 6.71 & \textbf{11.21} \\        \hline                                                                                                              \texttt{\textbf{LINK\_{uni}}} & 19.09 & 12.52 & 10.41 & 12.53 & 52.91 & 12.14 & \textbf{6.50} & 13.00 \\            \texttt{\textbf{LINK\_{domain}}} & \textbf{18.73} & 11.29 & \textbf{9.83} & 11.29 & \textbf{50.47} & 11.33 & 6.67 & 11.35 \\                                                                                       
\hline                                                                                                              \multicolumn{9}{c}{\cellcolor{gray!15}\textbf{1.3B}} \\                                                             \hline                                                                                                              LR (UB) & 7.33 & 21.42 & 5.17 & 21.68 & 9.12 & 40.76 & -- & -- \\                                                   \hdashline                                                                                                          LR & 58.12 & $>$10$^2$ & 37.44 & $>$10$^2$ & $>$10$^2$ & $>$10$^3$ & 11.87 & $>$10$^2$ \\                           LR + HR & 16.63 & \textbf{9.98} & 8.88 & \textbf{9.92} & 42.43 & \textbf{9.98} & 5.99 & \textbf{9.95} \\            \hline                                                                                                              \texttt{\textbf{LINK\_{uni}}} & 16.52 & 10.84 & 9.42 & 10.83 & 43.86 & 10.67 & \textbf{5.77} & 11.23 \\             \texttt{\textbf{LINK\_{domain}}} & \textbf{16.49} & 10.03 & \textbf{8.86} & 10.08 & \textbf{42.14} & 10.17 & 5.96 & 10.08 \\                                                                                       
\hline                                                                                                              \end{tabular}                                                                                                       \caption{Validation perplexity on FineWeb (EN) and FineWeb-2 (LR) across model sizes. Bold indicates lowest perplexity among LR~(S), LR+HR, \texttt{\textbf{LINK\_{uni}}}, and \texttt{\textbf{LINK\_{domain}}}.}  
\label{tab:app:ppl}                                                                                                 \end{table*}

\subsection{Global MMLU}

Global MMLU scores remain close to the random baseline of 25.0 across all setups and languages, with most values falling in the 25--28 range. 
At 1.3B scale, the models lack sufficient capacity to perform meaningfully on this knowledge-intensive benchmark. 
German and Chinese show the largest gains from bilingual training (up to 27.6), while low-resource languages remain near chance level regardless of the intervention strategy. 
These results suggest that Global MMLU is not sensitive enough to capture the differences between our setups at this model scale, and we therefore focus our analysis on the other benchmarks.

\begin{table}[h]
\centering
  \small
  \setlength{\tabcolsep}{3pt}
  \begin{tabular}{l|l|cccc|cccc}
  & & \multicolumn{4}{c|}{\textbf{High-resource}} & \multicolumn{4}{c}{\textbf{Low-resource}} \\
  \textbf{Size} & \textbf{Setup} & \textbf{DE} & \textbf{FR} & \textbf{ZH} & \textbf{HI} & \textbf{AM} & \textbf{IG} & \textbf{SW} & \textbf{YO} \\
  \hline
  \multirow{4}{*}{1.3B}
  & LR      & 25.8 & --   & 26.0 & 24.9 & 25.8 & 24.9 & 25.4 & 25.4 \\
  & LR + HR & 27.5 & 25.9 & 27.2 & 23.7 & 25.9 & 26.3 & 26.1 & 25.6 \\
  \cline{2-10}
  & \texttt{\textbf{LINK\_{uni}}} & 27.5 & 25.9 & 27.6 & 25.0 & 25.7 & 26.3 & 26.7 & 25.8 \\
  & \texttt{\textbf{LINK\_{domain}}}  & 27.4 & 27.3 & 27.6 & 23.8 & 25.5 & 26.7 & 26.6 & 25.7 \\
  \hline
  \multirow{4}{*}{345M}
  & LR      & 26.2 & 23.9 & 25.8 & 23.3 & 25.5 & 25.9 & 25.4 & 25.7 \\
  & LR + HR & 27.1 & 24.1 & 26.5 & 23.1 & 25.0 & 25.8 & 26.0 & 25.5 \\
  \cline{2-10}
  & \texttt{\textbf{LINK\_{uni}}} & 26.1 & 24.5 & 26.9 & 22.9 & 25.4 & 25.7 & 26.2 & 26.0 \\
  & \texttt{\textbf{LINK\_{domain}}}  & 26.4 & 26.0 & 26.6 & 22.9 & 25.1 & 25.9 & 25.6 & 25.6 \\
  \end{tabular}
  \caption{Global MMLU results in the target language across all model sizes. }
\label{tab:global_mmlu}
  \end{table}
  
\newpage
\subsection{Model Scaling}
\label{sec:app:models}

\begin{figure}[h!]
\centering
\includegraphics[width=16cm]{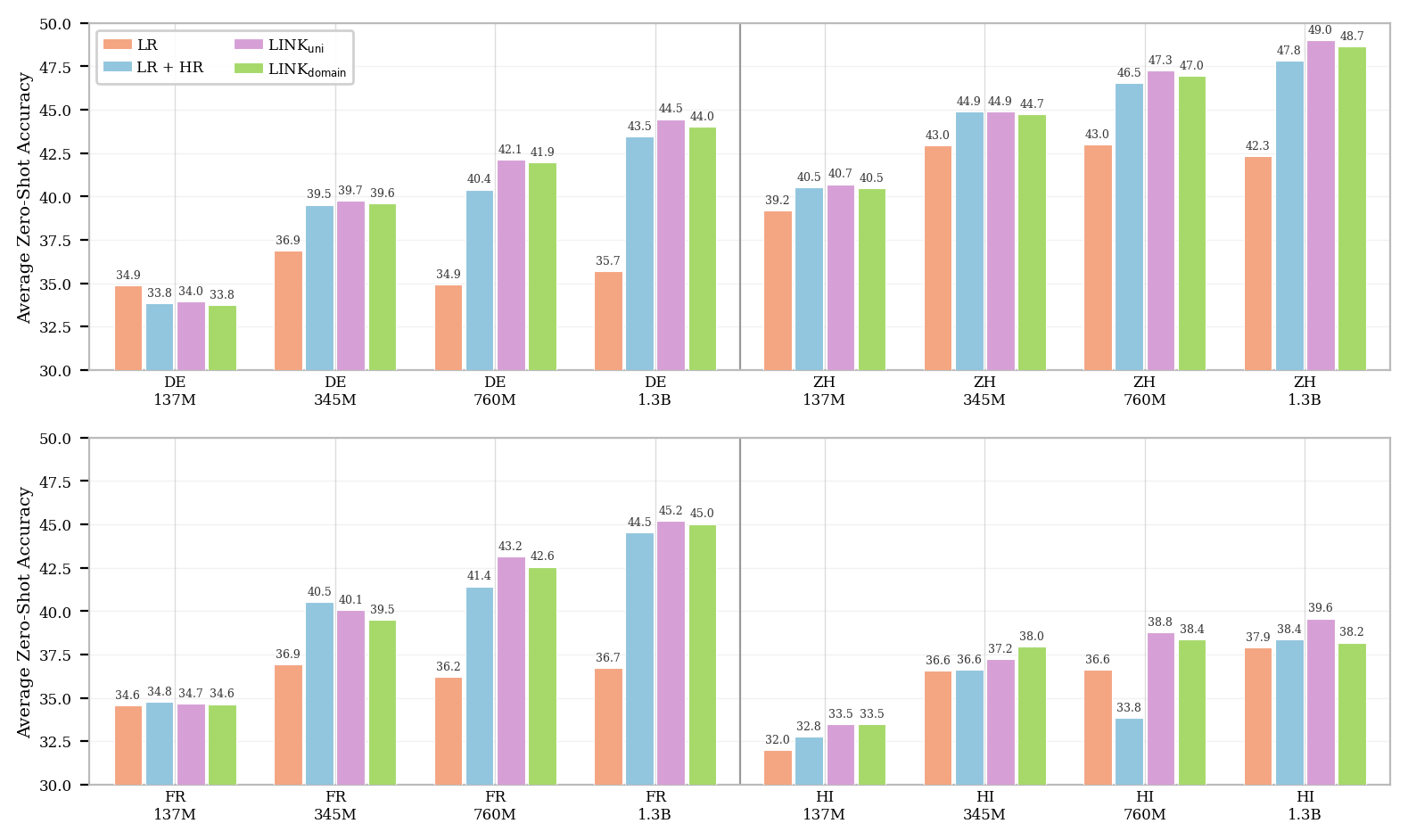}
\caption{Average benchmark accuracy across four model sizes and all languages.}
\label{fig:model_sizes}
\end{figure}

This section reports downstream benchmark results broken down by model size. 
Tables \ref{tab:app:760}, \ref{tab:app:345}, and \ref{tab:app:137} present individual benchmark scores for the 760M, 345M, and 137M models, respectively. 
Figure \ref{fig:model_sizes} summarizes the average zero-shot accuracy on target-language benchmarks across all four model sizes.

\begin{table*}[h!]
  \centering
  \small
  \setlength{\tabcolsep}{3pt}
  \begin{tabular}{l|ccccccc!{\vrule width 1pt}c||ccccccc!{\vrule width 1pt}c}
  &
  \multicolumn{8}{c||}{\textbf{LR performance}} &
  \multicolumn{8}{c}{\textbf{HR performance}} \\
  \cline{1-17}
  \textbf{Setup} &
  \textbf{A-E} & \textbf{A-C} & \textbf{HS} & \textbf{LB} & \textbf{PQ} & \textbf{SQ} & \textbf{WG} & \textbf{Avg} &
  \textbf{A-E} & \textbf{A-C} & \textbf{HS} & \textbf{LB} & \textbf{PQ} & \textbf{SQ} & \textbf{WG} & \textbf{Avg} \\
  \hline
  \multicolumn{17}{c}{\cellcolor{langDElight}\textbf{German}} \\
  \hline
  \rowcolor{langDEvlight}
  LR (UB) &
  34.2 & 26.1 & 37.9 & 26.7 & 62.4 & 55.2 & 50.4 & 41.8 & 32.0 & 22.2 & 30.5 & 24.6 & 58.2 & 53.1 & 50.5 & 38.7 \\
  \hdashline
  \rowcolor{langDElight}
  LR &
  31.5 & 23.4 & 28.5 & 9.1 & 56.0 & 45.5 & 50.4 & 34.9 & 27.5 & 24.5 & 26.7 & 2.6 & 49.9 & 40.8 & 49.2 & 31.6 \\
  \rowcolor{langDElight}
  LR + HR &
  34.7 & 24.4 & 32.6 & 21.8 & 55.3 & 62.0 & 51.9 & 40.4 & \textbf{49.5} & \textbf{26.8} & \textbf{49.4} & 45.4 & \textbf{71.1} & \textbf{72.9} & 54.3 & \textbf{52.8} \\
  \hline
  \rowcolor{langDE}
  \texttt{\textbf{LINK\_{uni}}} &
  36.4 & 24.1 & 34.1 & 23.6 & \textbf{59.1} & \textbf{64.6} & \textbf{53.0} & \textbf{42.1} & 44.6 & 24.5 & 46.1 & 43.2 & 69.4 & 68.4 & 52.9 & 49.9 \\
  \rowcolor{langDE}
  \texttt{\textbf{LINK\_{domain}}} &
  \textbf{36.8} & \textbf{25.5} & \textbf{35.2} & \textbf{23.7} & 58.8 & 63.0 & 50.7 & 41.9 & 47.4 & 26.4 & 49.4 & \textbf{46.1} & 71.1 & 72.9 & \textbf{54.5} & 52.5 \\
  \hline
  \multicolumn{17}{c}{\cellcolor{langFRlight}\textbf{French}} \\
  \hline
  \rowcolor{langFRvlight}
  LR (UB) &
  39.9 & 25.9 & 41.9 & 29.1 & 64.2 & 61.6 & 55.1 & 45.4 & 35.6 & 20.7 & 30.9 & 23.6 & 58.9 & 64.8 & 48.4 & 40.4 \\
  \hdashline
  \rowcolor{langFRlight}
  LR &
  31.4 & 22.9 & 30.1 & 10.7 & 54.8 & 51.8 & 51.9 & 36.2 & 28.5 & 21.7 & 27.4 & 4.7 & 53.1 & 45.8 & 50.9 & 33.2 \\
  \rowcolor{langFRlight}
  LR + HR &
  34.2 & 24.0 & 35.7 & 25.6 & 58.6 & 61.0 & 50.9 & 41.4 & \textbf{47.3} & \textbf{26.7} & 49.6 & \textbf{47.1} & 71.7 & \textbf{70.7} & \textbf{54.5} & \textbf{52.5} \\
  \hline
  \rowcolor{langFR}
  \texttt{\textbf{LINK\_{uni}}} &
  \textbf{37.4} & 25.2 & \textbf{36.9} & \textbf{27.0} & \textbf{60.3} & \textbf{63.2} & 52.3 & \textbf{43.2} & 43.9 & 24.5 & 45.6 & 42.8 & 69.4 & 69.6 & 54.4 & 50.0 \\
  \rowcolor{langFR}
  \texttt{\textbf{LINK\_{domain}}} &
  36.3 & \textbf{25.6} & 36.5 & 26.5 & 59.0 & 61.6 & \textbf{52.4} & 42.6 & 46.1 & 24.8 & \textbf{49.8} & 46.3 & \textbf{71.8} & 70.1 & 53.4 & 51.7 \\
  \hline
  \multicolumn{17}{c}{\cellcolor{langZHlight}\textbf{Chinese}} \\
  \hline
  \rowcolor{langZHvlight}
  LR (UB) &
  43.0 & 25.4 & 36.4 & -- & 60.1 & 76.7 & 51.4 & 48.8 & 31.9 & 20.5 & 27.9 & 12.1 & 54.0 & 59.3 & 48.9 & 36.4 \\
  \hdashline
  \rowcolor{langZHlight}
  LR &
  32.9 & 24.2 & 30.6 & -- & 55.6 & 63.5 & 51.2 & 43.0 & 26.9 & 23.1 & 26.7 & 1.6 & 52.5 & 39.5 & 51.1 & 31.6 \\
  \rowcolor{langZHlight}
  LR + HR &
  37.2 & 25.1 & 33.8 & -- & 57.0 & 74.0 & \textbf{52.1} & 46.5 & \textbf{48.0} & 25.4 & \textbf{49.8} & \textbf{45.6} & \textbf{72.4} & 71.7 & \textbf{53.8} & \textbf{52.4} \\
  \hline
  \rowcolor{langZH}
  \texttt{\textbf{LINK\_{uni}}} &
  \textbf{39.8} & \textbf{25.6} & \textbf{35.3} & -- & \textbf{57.7} & 73.7 & 51.6 & \textbf{47.3} & 47.0 & 25.2 & 46.7 & 42.2 & 69.6 & 70.9 & 52.6 & 50.6 \\
  \rowcolor{langZH}
  \texttt{\textbf{LINK\_{domain}}} &
  38.3 & 25.6 & 35.2 & -- & 56.9 & \textbf{75.3} & 50.5 & 47.0 & 47.4 & \textbf{26.4} & 49.3 & 45.2 & 70.8 & \textbf{71.8} & 52.8 & 51.9 \\
  \hline
  \multicolumn{17}{c}{\cellcolor{langHIlight}\textbf{Hindi}} \\
  \hline
  \rowcolor{langHIlight}
  LR &
  32.1 & 23.4 & 27.6 & -- & 53.0 & -- & 46.9 & 36.6 & 27.8 & 20.6 & 26.4 & 2.3 & 52.3 & 38.7 & 49.8 & 31.1 \\
  \rowcolor{langHIlight}
  LR + HR &
  32.4 & 22.3 & 28.7 & -- & 52.0 & -- & 49.1 & 36.9 & \textbf{48.4} & \textbf{26.4} & \textbf{49.8} & \textbf{46.8} & \textbf{71.6} & \textbf{73.1} & \textbf{52.6} & \textbf{52.7} \\
  \hline
  \rowcolor{langHI}
  \texttt{\textbf{LINK\_{uni}}} &
  33.0 & 24.1 & \textbf{30.1} & -- & \textbf{55.0} & -- & \textbf{51.6} & \textbf{38.8} & 43.8 & 23.8 & 44.6 & 40.5 & 68.4 & 69.2 & 52.1 & 48.9 \\
  \rowcolor{langHI}
  \texttt{\textbf{LINK\_{domain}}} &
  \textbf{33.5} & \textbf{24.5} & 29.7 & -- & 53.0 & -- & 51.2 & 38.4 & 46.5 & 25.2 & 49.0 & 46.0 & 71.2 & 69.7 & 51.2 & 51.3 \\
  \hline
  \end{tabular}
  \caption{760M model results.}
\label{tab:app:760}
\end{table*}

 \begin{table*}[h!]
  \centering
  \small
  \setlength{\tabcolsep}{3pt}
  \begin{tabular}{l|ccccccc!{\vrule width 1pt}c||ccccccc!{\vrule width 1pt}c}
  &
  \multicolumn{8}{c||}{\textbf{LR performance}} &
  \multicolumn{8}{c}{\textbf{HR performance}} \\
  \cline{1-17}
  \textbf{Setup} &
  \textbf{A-E} & \textbf{A-C} & \textbf{HS} & \textbf{LB} & \textbf{PQ} & \textbf{SQ} & \textbf{WG} & \textbf{Avg} &
  \textbf{A-E} & \textbf{A-C} & \textbf{HS} & \textbf{LB} & \textbf{PQ} & \textbf{SQ} & \textbf{WG} & \textbf{Avg} \\
  \hline
  \multicolumn{17}{c}{\cellcolor{langDElight}\textbf{German}} \\
  \hline
  \rowcolor{langDEvlight}
  LR (UB) &
  32.1 & 24.8 & 33.6 & 24.0 & 60.0 & 49.5 & 53.0 & 39.6 & 30.2 & 22.1 & 28.5 & 18.4 & 56.3 & 49.2 & 52.4 & 36.7 \\
  \hdashline
  \rowcolor{langDElight}
  LR &
  32.3 & 24.0 & 28.6 & 15.7 & 54.9 & 51.3 & 51.5 & 36.9 & 29.6 & 20.8 & 26.6 & 6.2 & 51.4 & 50.9 & 50.1 & 33.7 \\
  \rowcolor{langDElight}
  LR + HR &
  33.1 & 24.5 & 30.6 & \textbf{20.4} & 55.6 & 61.6 & 50.8 & 39.5 & 40.6 & 23.2 & 37.8 & 34.9 & 67.0 & 65.3 & 51.2 & 45.7 \\
  \hline
  \rowcolor{langDE}
  \texttt{\textbf{LINK\_{uni}}} &
  \textbf{33.6} & 23.5 & 30.5 & 19.4 & \textbf{55.8} & \textbf{61.7} & \textbf{53.6} & \textbf{39.7} & 41.5 & \textbf{24.2} & 38.0 & 34.6 & 66.2 & 66.1 & 50.4 & 45.9 \\
  \rowcolor{langDE}
  \texttt{\textbf{LINK\_{domain}}} &
  33.2 & \textbf{24.7} & \textbf{30.9} & 19.6 & 54.8 & 61.4 & 52.7 & 39.6 & \textbf{43.5} & 24.0 & \textbf{41.0} & \textbf{38.8} & \textbf{67.6} & \textbf{66.5} & \textbf{54.4} & \textbf{48.0} \\
  \hline
  \multicolumn{17}{c}{\cellcolor{langFRlight}\textbf{French}} \\
  \hline
  \rowcolor{langFRvlight}
  LR (UB) &
  35.5 & 24.5 & 37.1 & 23.9 & 62.2 & 61.8 & 52.7 & 42.5 & 33.1 & 20.6 & 28.7 & 16.9 & 56.1 & 58.1 & 50.2 & 37.7 \\
  \hdashline
  \rowcolor{langFRlight}
  LR &
  32.7 & 24.5 & 30.1 & 12.5 & 56.7 & 50.5 & 51.5 & 36.9 & 29.5 & 21.8 & 26.7 & 7.3 & 52.2 & 47.8 & 48.8 & 33.4 \\
  \rowcolor{langFRlight}
  LR + HR &
  \textbf{35.5} & 23.7 & 32.4 & \textbf{22.6} & \textbf{57.0} & 59.6 & \textbf{52.9} & \textbf{40.5} & 41.5 & 23.1 & 37.9 & 35.8 & 66.9 & 66.2 & 50.5 & 46.0 \\
  \hline
  \rowcolor{langFR}
  \texttt{\textbf{LINK\_{uni}}} &
  34.7 & \textbf{24.8} & \textbf{32.5} & 21.5 & 56.1 & \textbf{59.9} & 51.0 & 40.1 & 40.5 & 22.9 & 38.0 & 35.0 & 66.4 & 65.8 & 51.5 & 45.7 \\
  \rowcolor{langFR}
  \texttt{\textbf{LINK\_{domain}}} &
  33.3 & 24.3 & 32.0 & 22.3 & 56.5 & 58.0 & 49.9 & 39.5 & \textbf{43.9} & \textbf{24.1} & \textbf{40.8} & \textbf{38.1} & \textbf{67.6} & \textbf{68.5} & \textbf{51.7} & \textbf{47.8} \\
  \hline
  \multicolumn{17}{c}{\cellcolor{langZHlight}\textbf{Chinese}} \\
  \hline
  \rowcolor{langZHvlight}
  LR (UB) &
  39.7 & 25.1 & 33.5 & -- & 57.8 & 74.3 & 50.1 & 46.8 & 31.4 & 21.1 & 27.6 & 9.7 & 53.8 & 56.2 & 51.5 & 35.9 \\
  \hdashline
  \rowcolor{langZHlight}
  LR &
  31.4 & \textbf{25.0} & 30.2 & -- & 55.2 & 65.3 & \textbf{50.6} & 43.0 & 28.1 & 20.1 & 26.8 & 2.7 & 51.7 & 41.6 & 48.7 & 31.4 \\
  \rowcolor{langZHlight}
  LR + HR &
  36.5 & 24.0 & \textbf{32.4} & -- & 55.5 & \textbf{72.4} & 48.5 & 44.9 & 42.0 & 23.5 & 38.7 & 36.3 & 66.3 & \textbf{66.5} & 51.5 & 46.4 \\
  \hline
  \rowcolor{langZH}
  \texttt{\textbf{LINK\_{uni}}} &
  \textbf{36.6} & 24.2 & 32.1 & -- & \textbf{57.0} & 71.5 & 48.1 & \textbf{44.9} & 41.5 & \textbf{24.4} & 38.5 & 36.3 & 67.7 & 65.4 & 49.8 & 46.2 \\
  \rowcolor{langZH}
  \texttt{\textbf{LINK\_{domain}}} &
  35.9 & 24.4 & 32.2 & -- & 55.1 & 70.9 & 49.9 & 44.7 & \textbf{42.9} & 24.4 & \textbf{40.7} & \textbf{38.2} & \textbf{68.5} & 65.0 & \textbf{51.9} & \textbf{47.4} \\
  \hline
  \multicolumn{17}{c}{\cellcolor{langHIlight}\textbf{Hindi}} \\
  \hline
  \rowcolor{langHIlight}
  LR &
  31.0 & 23.1 & 27.9 & -- & 50.0 & -- & 50.9 & 36.6 & 29.6 & 20.3 & 26.2 & 3.4 & 52.7 & 42.2 & 49.7 & 32.0 \\
  \rowcolor{langHIlight}
  LR + HR &
  32.1 & 23.5 & \textbf{28.9} & -- & 47.0 & -- & \textbf{51.6} & 36.6 & 38.9 & \textbf{24.1} & 36.6 & 32.8 & 66.5 & 65.5 & 51.1 & 45.1 \\
  \hline
  \rowcolor{langHI}
  \texttt{\textbf{LINK\_{uni}}} &
  31.9 & \textbf{25.0} & 28.6 & -- & 49.0 & -- & 51.6 & 37.2 & 40.7 & 21.7 & 36.5 & 32.9 & 66.0 & 63.2 & 50.1 & 44.4 \\
  \rowcolor{langHI}
  \texttt{\textbf{LINK\_{domain}}} &
  \textbf{32.5} & 23.1 & 28.4 & -- & \textbf{56.0} & -- & 49.8 & \textbf{38.0} & \textbf{42.3} & 23.4 & \textbf{40.5} & \textbf{38.7} & \textbf{67.8} & \textbf{67.8} & \textbf{51.5} & \textbf{47.4} \\
  \hline
  \end{tabular}
  \caption{345M model results.}
\label{tab:app:345}
  \end{table*}

\begin{table*}[h!]
  \centering
  \small
  \setlength{\tabcolsep}{3pt}
  \begin{tabular}{l|ccccccc!{\vrule width 1pt}c||ccccccc!{\vrule width 1pt}c}
  &
  \multicolumn{8}{c||}{\textbf{LR performance}} &
  \multicolumn{8}{c}{\textbf{HR performance}} \\
  \cline{1-17}
  \textbf{Setup} &
  \textbf{A-E} & \textbf{A-C} & \textbf{HS} & \textbf{LB} & \textbf{PQ} & \textbf{SQ} & \textbf{WG} & \textbf{Avg} &
  \textbf{A-E} & \textbf{A-C} & \textbf{HS} & \textbf{LB} & \textbf{PQ} & \textbf{SQ} & \textbf{WG} & \textbf{Avg} \\
  \hline
  \multicolumn{17}{c}{\cellcolor{langDElight}\textbf{German}} \\
  \hline
  \rowcolor{langDEvlight}
  LR (UB) &
  27.2 & 23.6 & 26.6 & 12.2 & 53.5 & 41.9 & 50.0 & 33.6 & 28.5 & 21.5 & 27.0 & 5.4 & 52.0 & 41.4 & 49.6 & 32.2 \\
  \hdashline
  \rowcolor{langDElight}
  LR &
  \textbf{29.7} & \textbf{24.2} & 26.9 & \textbf{11.6} & \textbf{52.5} & 49.5 & 49.9 & \textbf{34.9} & 30.6 & \textbf{23.7} & 27.1 & 4.2 & 51.2 & 50.8 & \textbf{50.5} & 34.0 \\
  \rowcolor{langDElight}
  LR + HR &
  27.0 & 23.3 & \textbf{27.0} & 6.3 & 50.2 & \textbf{54.3} & 48.7 & 33.8 & \textbf{32.1} & 20.2 & \textbf{27.4} & \textbf{15.7} & \textbf{59.2} & \textbf{57.2} & 49.7 & \textbf{37.4} \\
  \hline
  \rowcolor{langDE}
  \texttt{\textbf{LINK\_{uni}}} &
  27.8 & 24.1 & 26.6 & 6.5 & 50.9 & 52.2 & 49.6 & 34.0 & 30.9 & 20.1 & 26.9 & 13.0 & 55.2 & 52.4 & 49.1 & 35.3 \\
  \rowcolor{langDE}
  \texttt{\textbf{LINK\_{domain}}} &
  28.2 & 21.7 & 26.5 & 6.5 & 49.7 & 52.7 & \textbf{50.9} & 33.8 & 31.9 & 20.0 & 27.4 & 14.4 & 57.5 & 53.7 & 47.9 & 36.1 \\
  \hline
  \multicolumn{17}{c}{\cellcolor{langFRlight}\textbf{French}} \\
  \hline
  \rowcolor{langFRvlight}
  LR (UB) &
  29.1 & 21.9 & 27.6 & 11.8 & 53.6 & 52.1 & 50.5 & 35.2 & 28.4 & 22.7 & 27.4 & 6.5 & 52.6 & 48.6 & 51.2 & 33.9 \\
  \hdashline
  \rowcolor{langFRlight}
  LR &
  28.4 & 23.1 & 27.2 & \textbf{10.9} & \textbf{52.5} & 49.5 & 50.4 & 34.6 & 29.0 & \textbf{22.8} & 27.1 & 4.2 & 52.2 & 46.7 & 51.9 & 33.4 \\
  \rowcolor{langFRlight}
  LR + HR &
  27.9 & 23.1 & 27.3 & 8.8 & 50.6 & 53.8 & \textbf{51.9} & 34.8 & \textbf{32.6} & 20.3 & 27.6 & \textbf{16.1} & \textbf{57.6} & \textbf{56.8} & 49.8 & \textbf{37.3} \\
  \hline
  \rowcolor{langFR}
  \texttt{\textbf{LINK\_{uni}}} &
  \textbf{28.5} & \textbf{23.9} & 27.2 & 9.7 & 51.7 & 52.5 & 49.1 & \textbf{34.8} & 30.8 & 20.9 & 26.9 & 13.7 & 55.1 & 51.7 & \textbf{51.9} & 35.9 \\
  \rowcolor{langFR}
  \texttt{\textbf{LINK\_{domain}}} &
  28.5 & 22.1 & \textbf{27.4} & 8.6 & 51.3 & \textbf{54.6} & 50.1 & 34.6 & 32.6 & 19.4 & \textbf{27.7} & 15.2 & 57.1 & 54.6 & 48.1 & 36.4 \\
  \hline
  \multicolumn{17}{c}{\cellcolor{langZHlight}\textbf{Chinese}} \\
  \hline
  \rowcolor{langZHvlight}
  LR (UB) &
  32.4 & 22.9 & 28.2 & -- & 52.6 & 65.8 & 51.3 & 42.2 & 29.1 & 21.4 & 27.0 & 3.5 & 52.7 & 45.3 & 49.4 & 32.6 \\
  \hdashline
  \rowcolor{langZHlight}
  LR &
  29.9 & 22.5 & \textbf{28.0} & -- & 50.7 & 52.8 & 51.4 & 39.2 & 28.4 & \textbf{24.0} & 26.8 & 1.8 & 50.8 & 38.5 & 48.7 & 31.3 \\
  \rowcolor{langZHlight}
  LR + HR &
  28.7 & 22.2 & 27.6 & -- & \textbf{52.3} & 61.9 & 50.5 & 40.5 & \textbf{33.1} & 20.6 & \textbf{27.6} & \textbf{16.3} & \textbf{58.8} & 53.3 & 48.9 & \textbf{37.0} \\
  \hline
  \rowcolor{langZH}
  \texttt{\textbf{LINK\_{uni}}} &
  28.8 & 22.8 & 27.2 & -- & 51.0 & 62.1 & \textbf{52.4} & \textbf{40.7} & 31.1 & 21.6 & 27.5 & 13.2 & 56.9 & 51.5 & \textbf{50.1} & 36.0 \\
  \rowcolor{langZH}
  \texttt{\textbf{LINK\_{domain}}} &
  \textbf{30.4} & \textbf{22.9} & 27.3 & -- & 51.0 & \textbf{62.4} & 48.9 & 40.5 & 31.6 & 21.2 & 27.3 & 14.8 & 57.7 & \textbf{53.8} & 49.7 & 36.6 \\
  \hline
  \multicolumn{17}{c}{\cellcolor{langHIlight}\textbf{Hindi}} \\
  \hline
  \rowcolor{langHIlight}
  LR &
  \textbf{29.8} & 22.4 & 26.8 & -- & 49.0 & -- & 49.6 & 35.5 & 28.3 & \textbf{22.6} & 26.4 & 2.9 & 52.5 & 40.5 & 49.9 & 31.9 \\
  \rowcolor{langHIlight}
  LR + HR &
  29.0 & 24.7 & 27.3 & -- & 50.0 & -- & 49.6 & 36.1 & \textbf{32.1} & 20.6 & \textbf{27.5} & \textbf{15.8} & \textbf{58.3} & \textbf{57.2} & 49.9 & \textbf{37.4} \\
  \hline
  \rowcolor{langHI}
  \texttt{\textbf{LINK\_{uni}}} &
  28.7 & \textbf{25.5} & 26.9 & -- & \textbf{53.0} & -- & 49.2 & 36.6 & 28.9 & 21.9 & 27.2 & 11.8 & 55.3 & 48.1 & \textbf{50.1} & 34.8 \\
  \rowcolor{langHI}
  \texttt{\textbf{LINK\_{domain}}} &
  29.2 & 24.2 & \textbf{27.4} & -- & 53.0 & -- & \textbf{50.0} & \textbf{36.8} & 31.9 & 19.9 & 27.5 & 15.3 & 58.1 & 53.7 & 47.4 & 36.2 \\
  \hline
  \end{tabular}
\caption{137M model results.}
\label{tab:app:137}
\end{table*}

Additionally, we conducted experiments on German for the 2.7B model; the results are provided in Table~\ref{tab:2_7B_results}. The gains of our methods are even more pronounced than for smaller models: for example, LINK\_{domain} achieves 45.2\% accuracy on ARC-Easy, surpassing not only all baselines but even the upper-bound model trained on significantly larger amounts of German data.

  \begin{table*}[t!]
  \centering
  \small
  \setlength{\tabcolsep}{3pt}
  \begin{tabular}{l|ccccccc!{\vrule width 1pt}c||ccccccc!{\vrule width 1pt}c}
  &
  \multicolumn{8}{c||}{\textbf{LR performance}} &
  \multicolumn{8}{c}{\textbf{HR performance}} \\
  \cline{1-17}

  \textbf{Setup} &
  \textbf{A-E} &
  \textbf{A-C} &
  \textbf{HS} &
  \textbf{LB} &
  \textbf{PQ} &
  \textbf{SQ} &
  \textbf{WG} &
  \textbf{Avg} &
  \textbf{A-E} &
  \textbf{A-C} &
  \textbf{HS} &
  \textbf{LB} &
  \textbf{PQ} &
  \textbf{SQ} &
  \textbf{WG} &
  \textbf{Avg}
  \\
  \hline
  \multicolumn{17}{c}{\cellcolor{langDElight}\textbf{German}} \\
  \hline
  \rowcolor{langDEvlight}
  LR (UB) &
  44.6 & 27.6 & 49.7 & 33.1 & 66.9 & 68.4 & 53.9 & 49.2 & 37.5 & 25.0 & 41.4 & 36.8 & 64.9 & 67.9 & 53.8 & 46.8 \\
  \hdashline
  \rowcolor{langDElight}
  LR &
  32.3 & 23.7 & 28.6 & 9.7 & 54.7 & 48.8 & 51.9 & 35.7 & 29.0 & 22.7 & 27.2 & 3.2 & 52.1 & 42.2 & 50.8 & 32.4 \\
  \rowcolor{langDElight}
  LR + HR &
  42.0 & \textbf{28.7} & 44.2 & 28.6 & 62.1 & 66.0 & 54.6 & 46.6 & 59.8 & \textbf{34.1} & 67.4 & 60.6 & 76.6 & \textbf{82.5} & \textbf{60.5} & 63.1 \\
  \hline
  \rowcolor{langDE}
  \texttt{\textbf{LINK\_{uni}}} &
  42.0 & 26.5 & 46.0 & \textbf{29.0} & 61.8 & \textbf{70.2} & \textbf{54.7} & 47.2 & 55.4 & 31.8 & 64.3 & 58.5 & 75.9 & 77.6 & 59.9 & 60.5 \\
  \rowcolor{langDE}
  \texttt{\textbf{LINK\_{domain}}} &
  \textbf{45.2} & 26.6 & \textbf{46.3} & 28.9 & \textbf{62.5} & 68.6 & 54.6 & \textbf{47.6} & \textbf{60.6} & 33.5 & \textbf{68.0} & \textbf{60.9} & \textbf{77.3} & 81.3 & 60.4 & \textbf{63.1} \\
  \hline
  \end{tabular}
  \caption{2.7B model results.}
  \vspace{-2pt}
  \label{tab:2_7B_results}
  \end{table*}

\newpage
\subsection{Full Low-Resource Results}
\label{sec:app:true_LRL}

This section reports downstream benchmark results for the true low-resource languages (Amharic, Igbo, Yoruba, and Swahili). 
The availability of target-language evaluation benchmarks varies by language: Amharic and Swahili have ARC-Easy, PiQA, and WinoGrande; Igbo has PiQA and WinoGrande; and Yoruba has ARC-Easy and PiQA translations.
The results are presented in Tables \ref{tab:app:lr_full_1_3} and \ref{tab:app:lr_full_345}.
Due to validation data limitation, we report the results for the last checkpoint.

\renewcommand{\arraystretch}{0.95}   

\begin{table*}[h!]
  \centering
  \small
  \setlength{\tabcolsep}{3pt}
  \renewcommand{\arraystretch}{1.1}
  \begin{tabular}{l|ccc!{\vrule width 1pt}c||ccccccc!{\vrule width 1pt}c}
  &
  \multicolumn{4}{c||}{\textbf{LR performance}} &
  \multicolumn{8}{c}{\textbf{HR performance}} \\
  \cline{1-13}

  \textbf{Setup} &
  \textbf{A-E} &
  \textbf{PQ} &
  \textbf{WG} &
  \textbf{Avg} &
  \textbf{A-E} &
  \textbf{A-C} &
  \textbf{HS} &
  \textbf{LB} &
  \textbf{PQ} &
  \textbf{SQ} &
  \textbf{WG} &
  \textbf{Avg}
  \\
  \hline
  \multicolumn{13}{c}{\textbf{Amharic}} \\
  \hline
  LR &
  \textbf{31.2} & 52.0 & \textbf{51.1} & \textbf{44.8} & 26.9 & 22.6 & 26.4 & 0.3 & 52.7 & 29.2 & 46.9 & 29.3 \\
  LR + HR &
  23.0 & \textbf{52.0} & \textbf{51.1} & 42.0 & \textbf{53.3} & \textbf{29.6} & \textbf{57.2} & \textbf{51.3} & \textbf{74.0} & \textbf{75.9} & \textbf{57.0} &
  \textbf{56.9} \\
  \hline
  \texttt{\textbf{LINK\_{uni}}} &
  24.6 & 51.0 & 49.1 & 41.6 & 51.5 & 29.7 & 56.0 & 51.1 & 74.0 & 73.8 & 55.2 & 55.9 \\
  \texttt{\textbf{LINK\_{domain}}} &
  25.9 & 47.0 & 50.0 & 41.0 & 52.5 & 27.7 & 57.1 & 52.8 & 74.4 & 76.9 & 54.6 & 56.6 \\
  \hline
  \multicolumn{13}{c}{\textbf{Igbo}} \\
  \hline
  LR &
  -- & 69.0 & 51.1 & 60.0 & 27.2 & 22.9 & 25.0 & 0.2 & 50.8 & 32.5 & 50.1 & 29.8 \\
  LR + HR &
  -- & \textbf{72.0} & \textbf{51.2} & \textbf{61.6} & \textbf{53.4} & \textbf{29.2} & \textbf{57.7} & \textbf{53.1} & 73.1 & \textbf{76.1} & \textbf{57.3} &
  \textbf{57.1} \\
  \hline
  \texttt{\textbf{LINK\_{uni}}} &
  -- & 71.0 & 49.7 & 60.4 & 52.6 & 28.2 & 57.5 & 52.2 & \textbf{74.9} & 76.1 & 56.0 & 56.8 \\
  \texttt{\textbf{LINK\_{domain}}} &
  -- & 71.0 & 49.7 & 60.4 & 53.6 & 28.2 & 57.6 & 52.2 & 73.7 & 75.9 & 56.7 & 56.8 \\
  \hline
  \multicolumn{13}{c}{\textbf{Yoruba}} \\
  \hline
  LR &
  \textbf{29.4} & 47.0 & -- & 38.2 & 25.8 & 21.8 & 25.9 & 0.1 & 51.3 & 28.0 & 50.7 & 29.1 \\
  LR + HR &
  24.9 & 55.0 & -- & 40.0 & 53.0 & 27.7 & 56.5 & 51.7 & 74.1 & 76.4 & 54.5 & 56.3 \\
  \hline
  \texttt{\textbf{LINK\_{uni}}} &
  24.5 & \textbf{57.0} & -- & \textbf{40.8} & 52.7 & \textbf{29.0} & 56.8 & 50.8 & 74.3 & \textbf{78.2} & 55.7 & \textbf{56.8} \\
  \texttt{\textbf{LINK\_{domain}}} &
  23.5 & 54.0 & -- & 38.8 & \textbf{53.0} & 28.2 & \textbf{56.1} & \textbf{52.6} & \textbf{73.4} & 75.7 & \textbf{56.8} & 56.5 \\
  \hline
  \multicolumn{13}{c}{\textbf{Swahili}} \\
  \hline
  LR &
  \textbf{32.2} & 61.0 & 50.3 & 47.8 & 30.6 & 21.6 & 26.5 & 2.2 & 52.6 & 40.0 & 50.0 & 31.9 \\
  LR + HR &
  28.1 & 67.0 & 50.5 & 48.5 & \textbf{55.0} & 29.0 & 57.5 & 52.9 & \textbf{74.6} & \textbf{78.5} & 56.0 & \textbf{57.6} \\
  \hline
  \texttt{\textbf{LINK\_{uni}}} &
  27.7 & 68.0 & 49.8 & 48.5 & 52.0 & 27.3 & 55.1 & 51.2 & 73.2 & 79.2 & 56.0 & 56.3 \\
  \texttt{\textbf{LINK\_{domain}}} &
  28.7 & \textbf{70.0} & \textbf{51.6} & \textbf{50.1} & 53.5 & \textbf{29.0} & \textbf{57.5} & \textbf{52.9} & 73.4 & 77.1 & \textbf{55.6} & 57.0 \\
  \hline
  \end{tabular}
  \caption{
  Results for low-resource languages (1.3B models). Results reported at the last checkpoint.
  A-E: ARC-Easy, PQ: PiQA, WG: WinoGrande.
  }
  \label{tab:app:lr_full_1_3}
  \end{table*}

\begin{table*}[h!]
  \centering
  \small
  \setlength{\tabcolsep}{3pt}
  \renewcommand{\arraystretch}{1.1}
  \begin{tabular}{l|ccc!{\vrule width 1pt}c||ccccccc!{\vrule width 1pt}c}
  &
  \multicolumn{4}{c||}{\textbf{LR performance}} &
  \multicolumn{8}{c}{\textbf{HR performance}} \\
  \cline{1-13}

  \textbf{Setup} &
  \textbf{A-E} &
  \textbf{PQ} &
  \textbf{WG} &
  \textbf{Avg} &
  \textbf{A-E} &
  \textbf{A-C} &
  \textbf{HS} &
  \textbf{LB} &
  \textbf{PQ} &
  \textbf{SQ} &
  \textbf{WG} &
  \textbf{Avg}
  \\
  \hline
  \multicolumn{13}{c}{\textbf{Amharic}} \\
  \hline
  LR &
  24.0 & 52.0 & \textbf{50.8} & 42.3 & 28.5 & 22.8 & 26.4 & 2.5 & 50.9 & 38.3 & 49.2 & 31.2 \\
  LR + HR &
  25.9 & 48.0 & 50.1 & 41.3 & \textbf{44.4} & 24.8 & \textbf{41.4} & 37.8 & 67.1 & 67.1 & \textbf{52.0} & \textbf{47.8} \\
  \hline
  \texttt{\textbf{LINK\_{uni}}} &
  \textbf{27.5} & \textbf{50.0} & 49.6 & \textbf{42.4} & 42.5 & 24.0 & 40.1 & 36.2 & \textbf{67.8} & 65.8 & 50.3 & 46.7 \\
  \texttt{\textbf{LINK\_{domain}}} &
  25.2 & \textbf{50.0} & 49.8 & 41.7 & 44.1 & 23.9 & 41.2 & \textbf{38.7} & 68.8 & \textbf{67.6} & 50.1 & \textbf{47.8} \\
  \hline
  \multicolumn{13}{c}{\textbf{Igbo}} \\
  \hline
  LR &
  -- & 58.0 & 48.5 & 53.2 & 27.3 & 24.1 & 25.9 & 0.7 & 49.9 & 37.1 & 49.7 & 30.7 \\
  LR + HR &
  -- & 65.0 & 50.0 & 57.5 & 43.1 & 24.0 & 41.3 & 38.6 & \textbf{68.8} & 64.9 & 51.3 & 47.4 \\
  \hline
  \texttt{\textbf{LINK\_{uni}}} &
  -- & \textbf{68.0} & \textbf{50.3} & \textbf{59.1} & 43.2 & 23.6 & 40.8 & 37.3 & 68.1 & \textbf{66.9} & \textbf{52.1} & 47.4 \\
  \texttt{\textbf{LINK\_{domain}}} &
  -- & 67.0 & 49.2 & 58.1 & \textbf{44.3} & \textbf{24.3} & \textbf{40.9} & \textbf{38.8} & 68.1 & 66.8 & 50.4 & \textbf{47.7} \\
  \hline
  \multicolumn{13}{c}{\textbf{Yoruba}} \\
  \hline
  LR &
  23.9 & \textbf{53.0} & -- & 38.5 & 25.9 & 20.9 & 25.1 & 0.2 & 49.6 & 29.7 & 49.3 & 28.7 \\
  LR + HR &
  22.7 & 47.0 & -- & 34.9 & \textbf{44.4} & 23.4 & 40.4 & 38.3 & 67.8 & 67.2 & \textbf{53.6} & \textbf{47.9} \\
  \hline
  \texttt{\textbf{LINK\_{uni}}} &
  \textbf{24.5} & \textbf{55.0} & -- & \textbf{39.8} & 41.8 & \textbf{24.6} & 40.4 & 37.2 & 67.8 & 66.7 & 51.5 & 47.1 \\
  \texttt{\textbf{LINK\_{domain}}} &
  22.5 & 44.0 & -- & 33.2 & 43.1 & \textbf{24.7} & \textbf{40.9} & \textbf{37.5} & \textbf{69.0} & \textbf{69.4} & 51.3 & \textbf{48.0} \\
  \hline
  \multicolumn{13}{c}{\textbf{Swahili}} \\
  \hline
  LR &
  \textbf{29.3} & 61.0 & \textbf{51.8} & 47.4 & 31.1 & 20.6 & 26.9 & 9.3 & 52.2 & 48.6 & 48.9 & 33.9 \\
  LR + HR &
  \textbf{29.3} & 61.0 & 49.0 & 46.4 & \textbf{44.1} & 22.9 & 41.0 & 38.1 & \textbf{69.4} & 68.1 & \textbf{52.6} & \textbf{48.0} \\
  \hline
  \texttt{\textbf{LINK\_{uni}}} &
  27.1 & \textbf{64.0} & 49.4 & 46.8 & 42.1 & 23.5 & 39.3 & 37.0 & 67.8 & 67.1 & 52.7 & 47.1 \\
  \texttt{\textbf{LINK\_{domain}}} &
  28.5 & \textbf{64.0} & 49.6 & \textbf{47.4} & 43.2 & \textbf{23.6} & \textbf{40.6} & \textbf{37.7} & 68.1 & 67.9 & 51.9 & 47.6 \\
  \hline
  \end{tabular}
  \caption{
  Results for low-resource languages (345M models). Results reported at the last checkpoint.
  A-E: ARC-Easy, PQ: PIQA, WG: WinoGrande.
  }
\label{tab:app:lr_full_345}
  \end{table*}

\begin{figure}[h!]
    \begin{center}
    \includegraphics[width=1.0\textwidth]{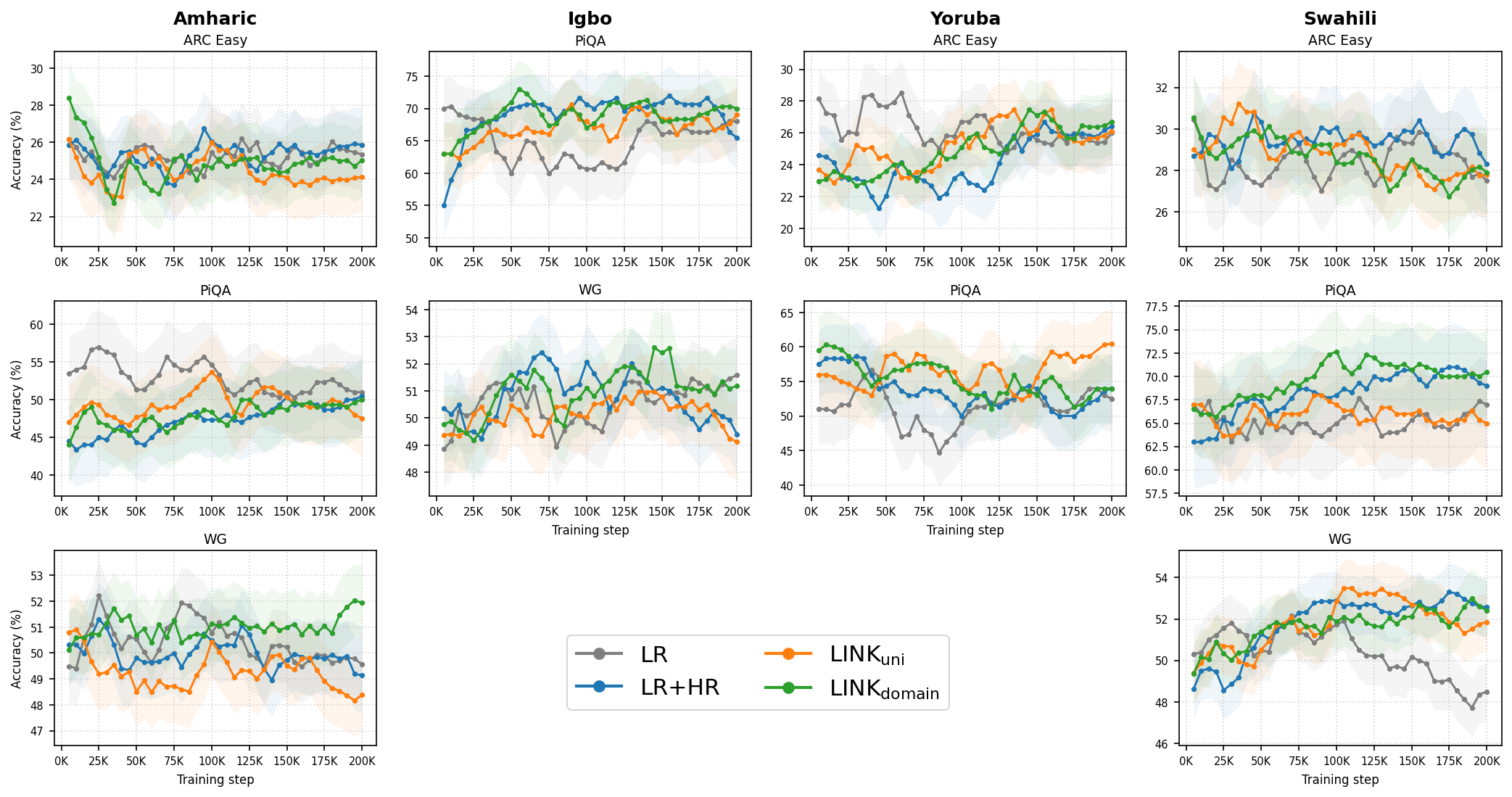}
    \end{center}
    \caption{1.3B models, extended (2$\times$) runs up to 200K steps: per-checkpoint accuracy on low-resource benchmarks (M-ARC Easy, M-PIQA, M-WinoGrande) for Amharic, Igbo, Yoruba, and Swahili. Shaded bands show standard deviation.}
    \label{fig:lrl_curves_1_3B}
\end{figure}

\begin{figure}[h!]
    \begin{center}
    \includegraphics[width=1.0\textwidth]{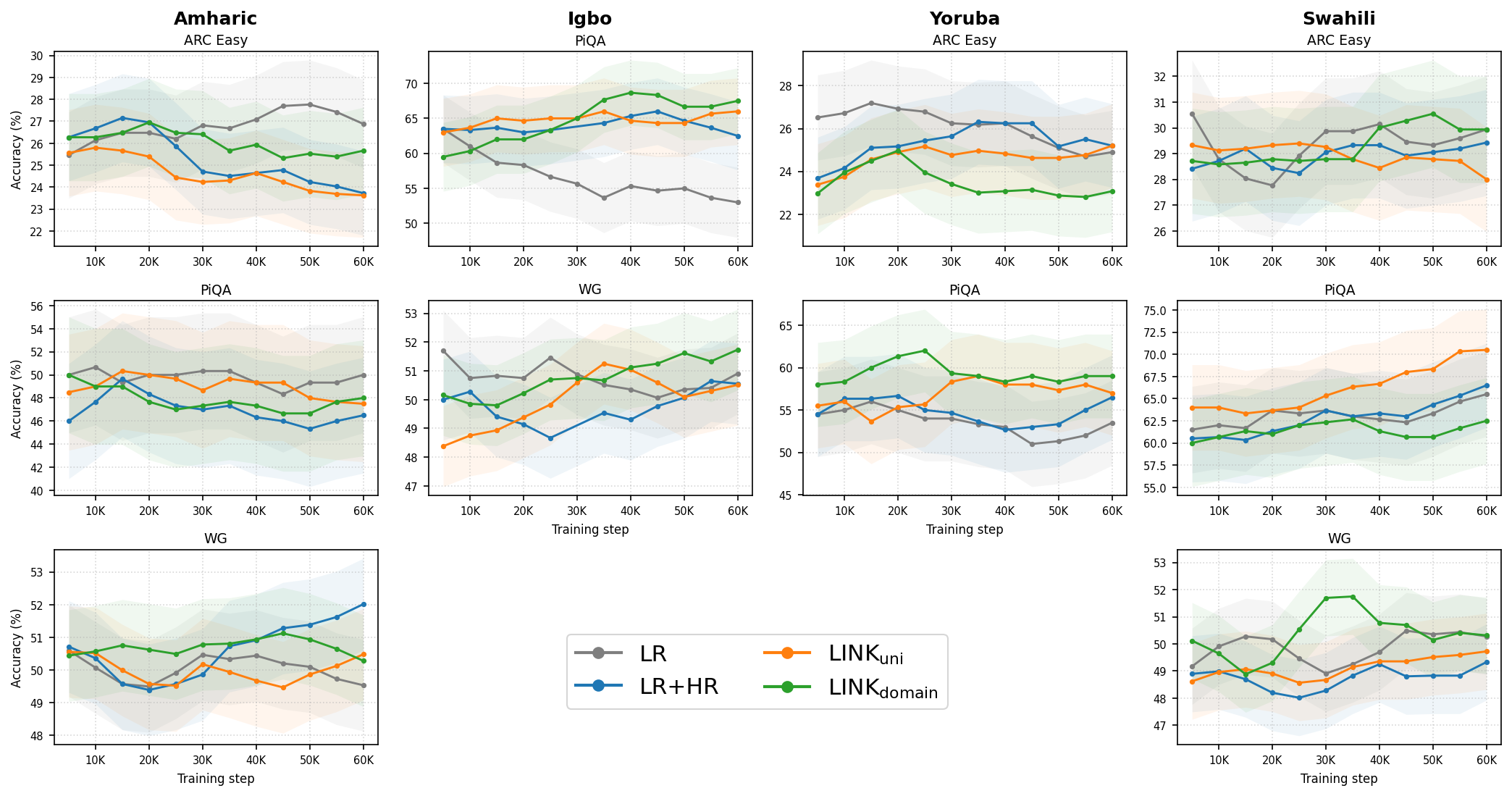}
    \end{center}
    \caption{345M models, extended (2$\times$) runs up to 60K steps: per-checkpoint accuracy on low-resource benchmarks (M-ARC Easy, M-PIQA, M-WinoGrande) for Amharic, Igbo, Yoruba, and Swahili. Shaded bands show standard deviation.}
    \label{fig:lrl_curves_345M}
\end{figure}

In order to better analyze the performance of our method for true low-resource languages, we ran an additional set of experiments with an extended number of training steps.
We kept the amount of available data the same as it was for the main experiments, but doubled the amount of training steps: i.e., the 1.3B models were trained for 200,000 steps, and 345M models were trained for 60,000 steps.
We report results across all checkpoints in Figures~\ref{fig:lrl_curves_1_3B} and \ref{fig:lrl_curves_345M}.

One issue with low-resource evaluations is benchmark noise, caused by the limited number available benchmarks for these languages as well as their small size.
For example, the GlobalPiQA benchmark~\citep{Chang2025GlobalPE} we used for this evaluation contains only 100 samples per language.
While target-language benchmarks for high-resource languages such as German exhibit low variance (e.g., 0.6--1.1pp between consecutive checkpoints), low-resource benchmarks are 2--3$\times$ noisier: Igbo PiQA fluctuates by up to 16 points between consecutive checkpoints, Yoruba PiQA by up to 10 points, and Amharic PiQA by up to 6 points.

\newpage
\subsection{Clustering Results}
\label{sec:app:clustering}

\begin{figure}[h]
    \begin{center}
    \includegraphics[width=0.8\textwidth]{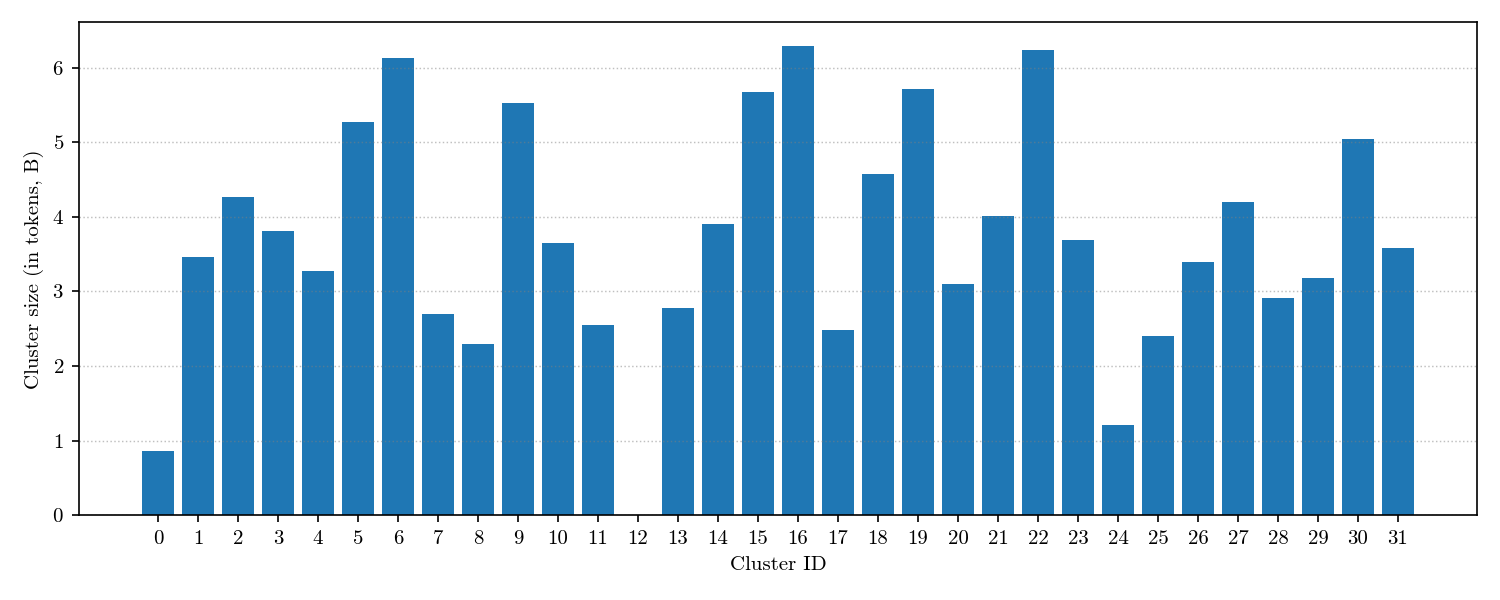}
    \end{center}
    \caption{Distribution of English FineWeb training tokens across the 32 k-means clusters.}
    \label{fig:fineweb_clusters}
\end{figure}

Figure \ref{fig:fineweb_clusters} demonstrates the results of the k-means clustering (32 clusters) of the English FineWeb training data. 
Most clusters contain around 2-6B tokens apart from cluster~12 which is almost empty (0.21M tokens, i.e.,~${\sim}$0.0002\% of all 118B total tokens). 
Cluster~5 hold around 5.3B tokens (5.55M samples), which corresponds to 4.46\% tokens (3.76\% samples) of the whole Fineweb dataset.

\begin{figure}[h]
    \begin{center}
    \includegraphics[width=0.8\textwidth]{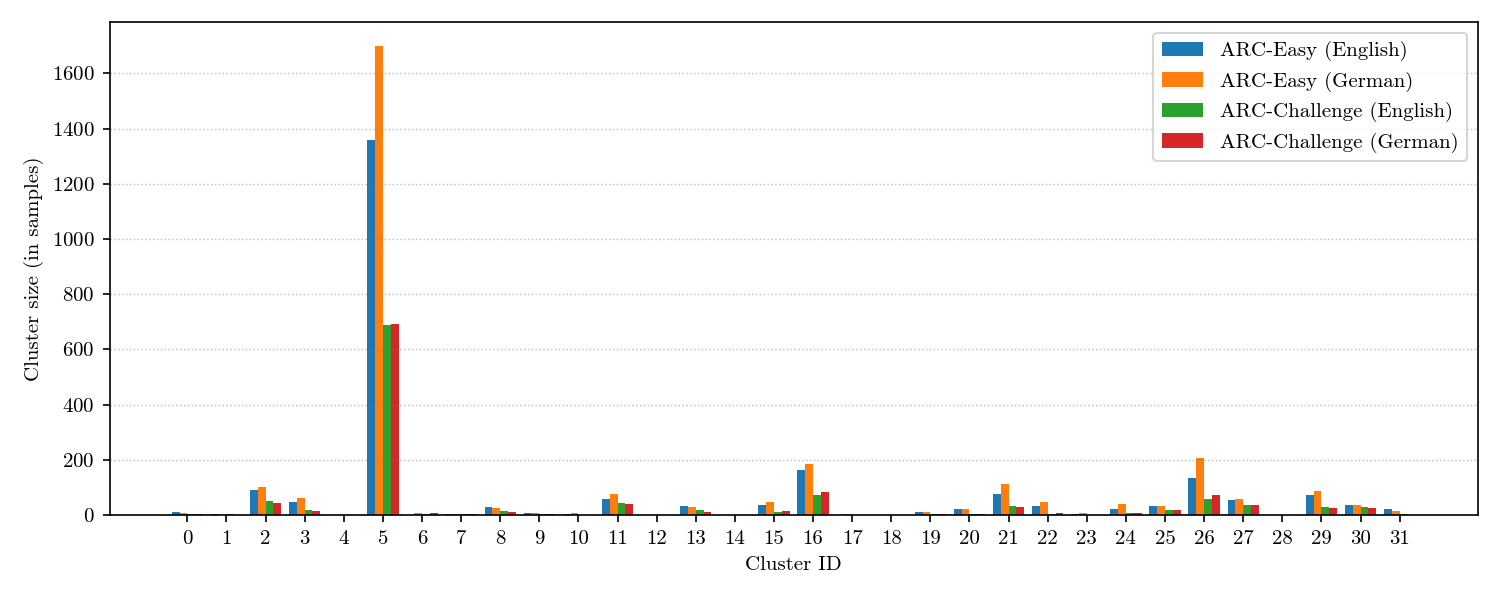}
    \end{center}
    \caption{Results of clustering of ARC-Easy and ARC-Challenge validation datasets.}
    \label{fig:clusters}
\end{figure}

Next, we clustered ARC-Easy and ARC-Challenge benchmarks (both English and German versions) using the same centroids. The results are provided in Figure \ref{fig:clusters}.  

Both benchmarks are heavily concentrated in a single Cluster 5, which accounts for more than half of the samples of the corresponding benchmarks: 57.24\% of samples from ARC-Easy (English), 57.67\% of samples from ARC-Easy (German), 58.87\%  of samples from ARC-Challenge (English), 59.11\% of samples from ARC-Challenge(German).
The manual inspection of the resulted cluster further confirm the initial hypothesis of scientific knowledge (which is mostly represented in ARC datasets) being concentrated in one cluster, what allows us to use it for  {\texttt{LINK\_{domain}}} experiments.
\subsection{Target vs Actual Replacements}
\label{sec:app:target_actual_repl} 

\begin{figure}[h!]
    \begin{center}
    \includegraphics[width=1.0\textwidth]{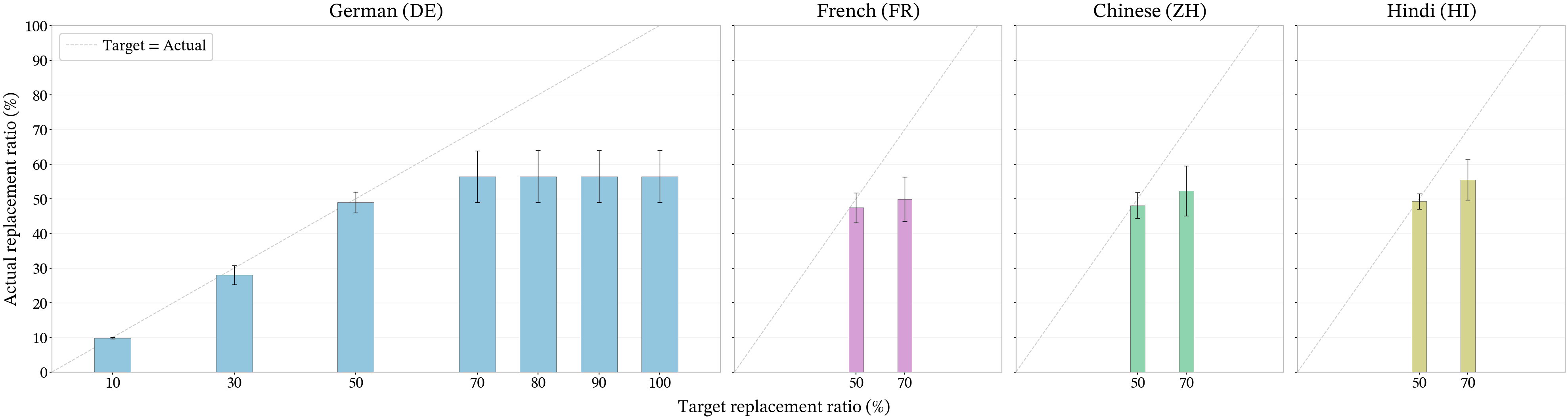}
    \end{center}
    \caption{Target vs. actual replacement ratio for each language. The dashed line indicates the ideal case where all targeted words are replaced. The gap between target and actual ratios reflects bilingual vocabulary coverage — languages with smaller vocabularies (e.g., Hindi) saturate at lower actual replacement rates.}
    \label{fig:target_act_repl}
\end{figure}

Figure \ref{fig:target_act_repl} shows the relationship between the target and actual replacement ratios for each language. 
Since replacements can only be performed for words present in the bilingual vocabulary, the actual replacement ratio is bounded by vocabulary coverage. 
For German, which has the largest vocabulary (48,195 entries), the actual ratio closely tracks the target up to 30\%, after which it plateaus around 55–57\%. We conducted an extensive ablation over replacement ratios for German (Figure \ref{fig:target_act_repl}, left), and based on the finding that actual replacements saturate beyond a target ratio of 70, we evaluated only the 50 and 70 settings for the remaining languages.

French, Chinese, and Hindi exhibit a similar ceiling effect at target ratios of 50 and 70, with actual ratios reaching approximately 47–50\%, 48–55\%, and 49–55\% respectively. 
The gap between the target and actual ratios varies across languages, reflecting differences in vocabulary size. 
Based on these results, we set the target replacement ratio to 70 for all experiments, as this effectively maximizes the number of replacements achievable with our vocabularies.

\begin{figure}[h!]
    \begin{center}
    \includegraphics[width=0.7\textwidth]{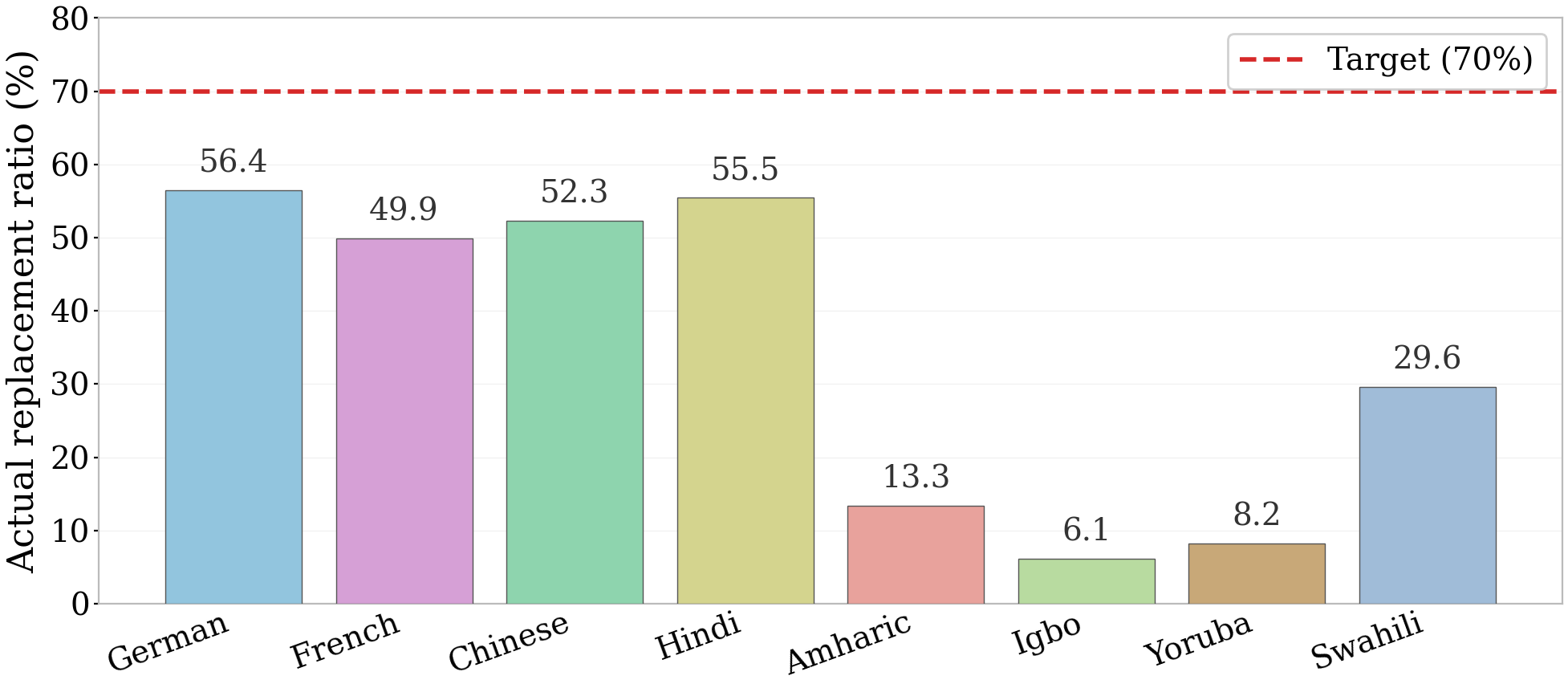}
    \end{center}
    \caption{Actual per-document replacement rate at target=70\% for all eight languages. The dashed red line marks the 70\% target. High-resource languages saturate around 50--56\%; truly low-resource languages fall far below the target, bounded by the size of the available bilingual vocabularies.}
    \label{fig:actual_repl_70_all_langs}
\end{figure}

The vocabulary-coverage ceiling is far more pronounced for truly low-resource languages. Figure~\ref{fig:actual_repl_70_all_langs} reports the actual per-document replacement rate at the fixed target of 70\% across all eight languages used in this work. The four data-constrained high-resource languages reach the saturation level discussed above (50–56\%), whereas the four truly low-resource languages fall dramatically below the target: Swahili reaches only 29.6\%, Amharic 13.3\%, Yoruba 8.2\%, and Igbo 6.1\%. These rates are upper-bounded by the size of the available bilingual vocabularies (Table~\ref{tab:data_stat}); even when every replaceable word is replaced, the per-document rate cannot exceed the fraction of words covered by the dictionary. 

\subsection{Reduced Vocabulary Size Experiments}
\label{sec:app:vocab_reduction}

\begin{wrapfigure}{r}{0.5\textwidth}
\vspace{-10pt}
\centering
\includegraphics[width=0.5\textwidth]{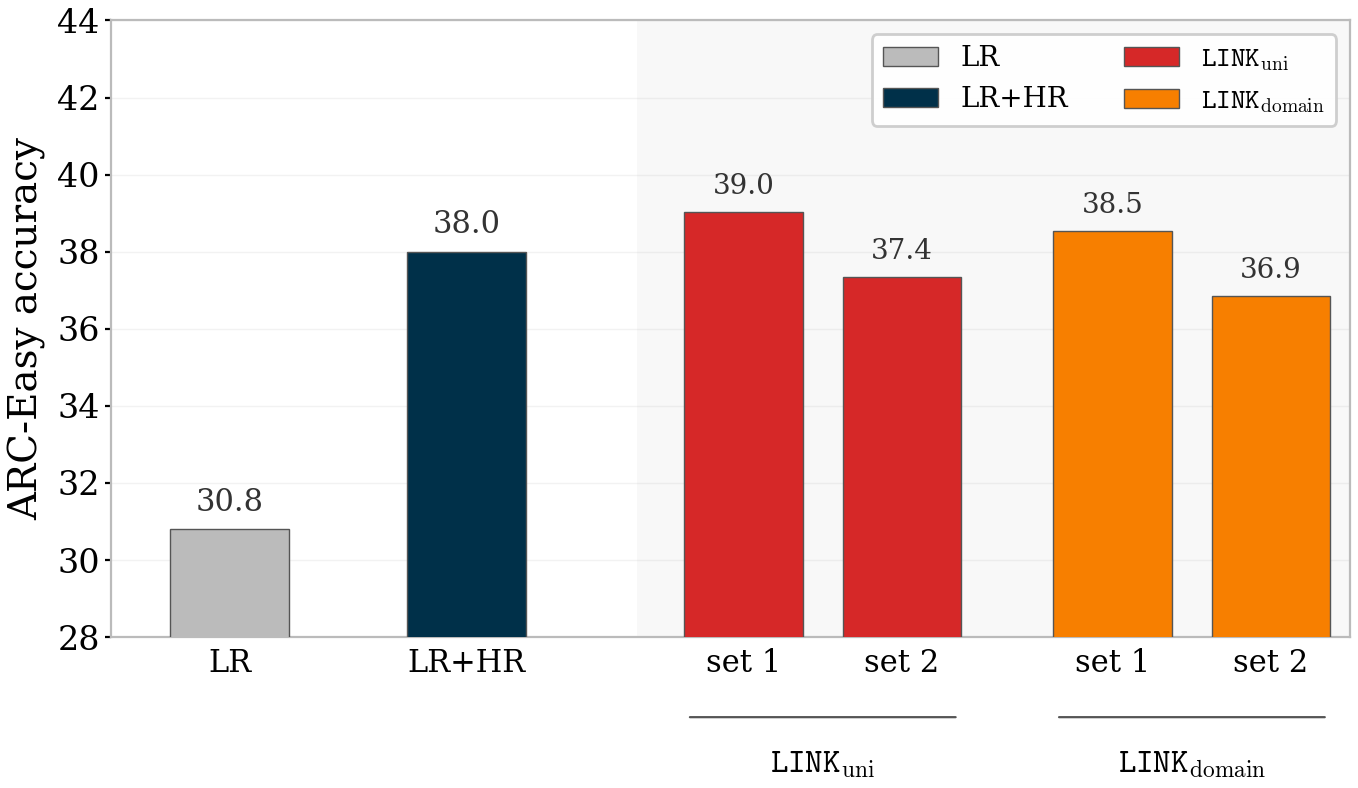}
\caption{ARC-Easy accuracy on German at 1.3B with bilingual vocabularies reduced to 1\% of the original size (around 480 word pairs). Set 1 and Set 2 are two independently subsampled 1\% vocabularies (different sampling seeds).}
\vspace{-10pt}
\label{fig:reduced_vocab_1pct}
\end{wrapfigure}
In addition to the experiments discussed in Section \ref{sec:ablation}, we further reduced the original German vocabulary to 1\% of the initial size - i.e., to around 480 word pairs.
This vocabulary was subsampled from 10\% vocabulary with two different seeds.
The results presented in Figure \ref{fig:reduced_vocab_1pct} demonstrate that when the bilingual vocabulary is small, it becomes increasingly important which word pairs it contains (not only how many).
Across the two independently sampled 1\% vocabularies, ARC-Easy accuracy varies by 1.5-1.7pp for both \texttt{LINK\_{uni}} (39.0 vs.\ 37.4) and \texttt{LINK\_{domain}} (38.5 vs.\ 36.9), indicating that vocabulary composition becomes a primary driver of transfer at this scale.
The stronger of the two subsets still matches or exceeds the LR+HR baseline (38.0), while the weaker one falls slightly below it, implying that with only a few hundred translation pairs available, careful selection of entries matters.

\subsection{Data-Constrained vs Low-Resource}
\label{sec:app:dc_lr}

Throughout this work, we use the term \textit{data-constrained} rather than \textit{low-resource} to describe our experimental settings. While the two terms refer to overlapping concepts, they highlight distinct challenges. Our method is broadly applicable to any language for which training data is limited, regardless of whether it is traditionally classified as low-resource.

Several factors motivate this distinction.
First, truly low-resource languages typically have limited bilingual vocabulary coverage, whereas our simulated settings use languages with large bilingual dictionaries. 
Second, our simulated settings are constructed by downsampling web-crawled data, which preserves the topical diversity of the original corpus. 
In practice, low-resource language data is often drawn from a narrow set of sources (religious texts, government documents, or Wikipedia) resulting in a skewed domain distribution that our downsampling procedure does not capture.
Third, low-resource languages frequently lack standardized evaluation benchmarks, limiting our ability to assess model performance comprehensively. 

To avoid conflating these distinct challenges, we reserve the term \textit{low-resource} for the experiments in Section~\ref{sec:true_LRL}, supporting truly low-resource languages remains a core motivation of this work, and use \textit{data-constrained} elsewhere.

\applefootnote{ \textcolor{textgray}{\sffamily Apple and the Apple logo are trademarks of Apple Inc., registered in the U.S. and other countries and regions.}}

\end{document}